\documentclass[]{article} % For LaTeX2e
\usepackage{fullpage}
\usepackage{url}
\usepackage{color}
\usepackage{cancel}
\usepackage{iclr2016_conference,times}
\usepackage{url}

 \iclrfinaltrue
\usepackage{amsmath,amssymb}
\usepackage{amsthm}
\usepackage{graphicx}
\usepackage[tight,scriptsize]{subfigure}
\usepackage{enumitem}

\def\E{\mathbb{E}}
\def\real{\mathbb{R}}
\def\eg{{\em e.g.,~}}
\def\ie{{\em i.e.,~}}
\def\cutForReview#1{}

\newcommand{\be}{\begin{equation}}
\newcommand{\ee}{\end{equation}}
\newcommand{\bea}{\begin{eqnarray}}
\newcommand{\eea}{\end{eqnarray}}

\newcommand{\vspacetitle}{\vspace{-0cm}}

\newtheorem{claim}{Claim} 
\newtheorem{rem}{Remark} 
\newtheorem{thm}{Theorem} 

\newtheorem{lemma}{Lemma}
\newtheorem{defn}{Definition}
\newtheorem{cor}{Corollary}

\def\xx{{I}}
\def\M{{\xi}}

\def\xx{{\bf y}}
\def\M{{\bf z}}

\def\T{{T}}
\def\ttheta{{\alpha}}

\def\tt{{\tau}}

\def\xx{{x}}
\def\M{{\theta}} % CAREFUL IT INTERFERES WITH \ttheta, gradient orientation

\def\XX{{X}}

\def\T{{T}}

\def\f{{y}}

\usepackage[font={footnotesize}]{caption}

\usepackage[pagebackref=true,breaklinks=true,letterpaper=true,colorlinks,bookmarks=false]{hyperref}

\title{Visual Representations: \\ {\large Defining Properties and Deep Approximations}} 

\author{Stefano Soatto  \\
Department of Computer Science\\
University of California, Los Angeles\\
Los Angeles, CA 90095, USA \\
\texttt{soatto@ucla.edu} \\
\AND
Alessandro Chiuso \\
Dipartimento di Ingegneria dell'Informazione \\ 
Universit\`a di Padova \\
Via Gradenigo 6/b, 35131 Padova, Italy \\
\texttt{alessandro.chiuso@unipd.it} 
}

\begin{document}

\maketitle

\begin{abstract} 
Visual representations are defined in terms of minimal sufficient statistics of visual data, for a class of tasks, that are also invariant to nuisance variability. Minimal sufficiency guarantees that we can store a representation in lieu of raw data with smallest complexity and no performance loss on the task at hand. Invariance guarantees that the statistic is constant with respect to uninformative transformations of the data. We derive analytical expressions for such representations and show they are related to feature descriptors commonly used in computer vision, as well as to convolutional neural networks. This link highlights the assumptions and approximations tacitly assumed by these methods and explains empirical practices such as clamping, pooling and joint normalization. 
\end{abstract}

\section{Introduction}
\vspacetitle

A {\em visual representation} is a function of visual data (images) that is ``useful'' to accomplish visual tasks. Visual tasks are decision or control actions concerning the surrounding environment, or {\em scene,} and its properties.  Such properties can be geometric (shape, pose), photometric (reflectance), dynamic (motion) and semantic (identities, relations of ``objects'' within). In addition to such properties, the data also depends on a variety of {\em nuisance variables} that are irrelevant to the task. Depending on the task, they may include unknown characteristics of the sensor (intrinsic calibration), its inter-play with the scene (viewpoint, partial occlusion), and properties of the scene that are not directly of interest (\eg illumination). 

We are interested in modeling and analyzing visual representations: How can we measure how ``useful'' one is? What guidelines or principles should inform its design? Is there such a thing as an {\em optimal} representation? If so, can it be computed? Approximated? Learned?
 We abstract (classes of) visual tasks to {\em questions about the scene}. They could be countable semantic queries (\eg concerning ``objects'' in the scene and their relations) or continuous control actions (\eg ``in which direction to move next'').

In Sect. \ref{sect-desiderata} we formalize these questions using well-known concepts, and in Sect. \ref{sect-lik} we derive an equivalent characterization that will be the starting point for designing, analyzing and learning representations.

%%%%%%%%%%%%%%%%%%%%%%%%%%%%%%%%%%%%%%%%%%%%%%%%%%
%									   Related Work
%%%%%%%%%%%%%%%%%%%%%%%%%%%%%%%%%%%%%%%%%%%%%%%%%%
\vspacetitle
\subsection{Related Work and Contributions}
\label{sect-related}
\vspacetitle

Much of Computer Vision is about computing functions of images that are ``useful'' for visual tasks. When restricted to subsets of images, {\em local descriptors} typically involve statistics of image gradients, computed at various scales and locations, pooled over spatial regions, variously normalized and quantized.  This process is repeated hierarchically in a convolutional neural network (CNN), with weights inferred from data, leading to {\em representation learning} \cite{ranzato2007unsupervised,lecun2012learning,simonyan2014learning,serre2007feedforward,rosasco,susskind,bengio2009learning}. 
There are more methods than we can review here, and empirical comparisons (\eg \cite{mikolajczyk04comparison}) have recently expanded to include CNNs. Unfortunately, many implementation details and parameters make it hard to draw general conclusions \cite{chatfield2011devil}.
We take a different approach, and derive a formal expression for optimal representations from established principles of {\em sufficiency, minimality, invariance.} We show how existing descriptors are {\em related} to such representations, highlighting the (often tacit) underlying assumptions. 

Our work relates most closely to \cite{mallatB11,anselmi2015invariance} in its aim to construct and analyze representations for classification tasks. However, we wish to represent the {\em scene}, rather than the image, so occlusion and locality play a key role, as we discuss in Sect. \ref{sect-discussion}. Also \cite{morel2011sift} characterize the invariants to certain nuisance transformations: Our models are more general, involving both the range and the domain of the data, although more restrictive than \cite{tishbyPB00} and specific to visual data. We present an alternate interpretation of pooling, in the context of classical sampling theory, that differs from other analyses \cite{gong2014multi,boureau2010theoretical}. 

Our  contributions are to (i) define minimal sufficient invariant representations and characterize them explicitly (Claim \ref{claim-sal}); (ii) show that local descriptors currently in use in Computer Vision can approximate  such representations under very restrictive conditions (Claim \ref{claim-dsp} and Sec. \ref{sect-contrast}); (iii) compute in closed form the minimal sufficient {\em contrast} invariant \eqref{eq-contr-inv} and show how local descriptors relate to it (Rem. \ref{rem-sift}); show that such local descriptors can be implemented via linear convolutions and rectified linear units (Sect. \ref{sect-layer1}); (iv) explain the practice of ``joint normalization'' (Rem. \ref{rem-joint}) and ``clamping'' (Sect. \ref{sect-clamping}) as procedures to approximate the sufficient invariant; these practices are seldom explained and yet they have a significant impact on performance in empirical tests  \cite{kirchner15}; (v) explain ``spatial pooling'' in terms of {\em anti-aliasing,} or local marginalization with respect to a small-dimensional nuisance group, in convolutional architectures (Sec. \ref{sect-stacking-simplifies}-\ref{sect-local}). In the Appendix we show that an ideal representation, if generatively trained, maximizes the information content of the representation (App. \ref{sect-likelihood}). 

\vspacetitle
\section{Characterization and properties of representations} 
\label{sect-characterization}
\vspacetitle

Because of uncertainty in the mechanisms that generate images, we treat them as realizations of random vectors $\xx$ (past/training set) and $\f$ (future/test set), of high but finite dimension. The scene they portray is infinitely more complex.\footnote{Scenes are made of surfaces supporting reflectance functions that interact with illumination, etc. No matter how many images we already have, even a single static scene can generate infinitely many different ones.} Even if we restrict it to be a static ``model'' or ``parameter'' $\M$, it is in general infinite-dimensional. Nevertheless, we can {\em ask questions} about it. The number of questions (``classes'') $K$ is large but finite, corresponding to a partition of the space of scenes,  represented by samples $\{\M_1, \dots, \M_K\}$. A simple model of image formation including scaling and occlusion \cite{soattoD12ICVSS} is known as the Lambert-Ambient, or LA, model.

\vspacetitle
\subsection{Desiderata} 
\label{sect-desiderata}
\vspacetitle

Among (measurable) functions of past data (statistics), we seek those useful for a class of tasks. Abstracting the task to {\em questions} about the scene, ``useful'' can be measured by uncertainty reduction on the answers, captured by the {\em mutual information} between the data $\xx$ and the object of interest $\M$.  While a representation can be no more informative than the data,\footnote{Data Processing Inequality, page 88 of \cite{shao98}.} ideally it should be no less, \ie a {\em sufficient statistic}. It should also be ``simpler'' than the data itself, ideally {\em minimal}. It should also discount the effects of nuisance variables $g \in G$, and ideally be {\em invariant} to them. We use a superscript to denote a collection of $t$ data points (the history up to $t$, if ordered), $\xx^t \doteq \{\xx_1, \dots, \xx_t\}$.
\begin{quote}
Thus, {\em a representation is any function $\phi$ constructed using past data $\xx^t$ that is useful to answer questions about the {\em scene} $\M$ given future data $\f$ it generates, regardless of nuisance factors $g$.}  
\end{quote}
An optimal representation is  {\em a minimal sufficient statistic for a task that is invariant to nuisance factors.} In Sec. \ref{sect-lik} we introduce the SA Likelihood, an optimal representation, and in subsequent sections show how it can be approximated.

\vspacetitle
\subsection{Background}
\label{sect-background}
\vspacetitle

The data $\XX$ is a random variable with samples $\xx, \f$; the model $\M$ is unknown in the experiment $E = \{\xx, \M, p_\M(\xx)\}$ where $p_\M(\xx)$ is the probability density function of $\XX$,  that depends on the parameter $\M$, evaluated at the sample $\xx$; a {\em statistic} $\T$ is a function of the sample; it is {\em sufficient} (of $\xx$ for $\M$) if  $\XX \ | \ \T = \tt$ does not depend on $\M$;\footnote{Definition 3.1 of \cite{pawitan} or Sec. 6.7 of \cite{degroot89}, page 356} it is {\em minimal} if it is a function of all other sufficient statistics\footnote{If $U$ is sufficient, then the sigma algebra $\sigma(T) \subset \sigma(U)$, \cite{degroot89}, page 368.}. If $\M$ is treated as a random variable and a prior is available, $\phi$ is Bayesian sufficient\footnote{The two are equivalent for discrete random variables, but pathological cases can arise in infinite dimensions \cite{blackwell1982bayes}.} if $p(\M | \phi(\xx^t)) = p(\M | \xx^t)$. 

If $\T$ is minimal, any smaller\footnote{In the sense of inclusion of sigma algebras, $\sigma(U) \subset \sigma(T)$.} $U$ entails ``information loss.''  
If $\M$ was a discrete random variable, the information content of $T$ could be measured by uncertainty reduction: ${\mathbb H}(\M) - {\mathbb H}(\M | \T(\XX))$, which is the mutual information\footnote{See \cite{coverT} eq. 2.28, page 18.}  between $\M$ and $T$ and ${\mathbb H}$ denotes entropy \cite{coverT}; furthermore, $T(\XX) \in \arg\inf_{\phi} {\mathbb H}(\M | \phi(\XX))$, where the infimum is with respect to measurable functions and is in general not unique. 

Consider a set $G$ of transformations $g$ of the data $\xx$, $g(\xx)$, which we denote simply as $g\xx$.  A function $\phi_{{}_G}(\xx)$ is $G$-invariant if $\phi_{{}_G}(g\xx) = \phi_{{}_G}(\xx)$ for all $g \in G$.  The {\em sensitivity} of $\phi$ to $G$ is $S = \| \frac{\partial \phi(g\xx)}{\partial g} \|$ where $\phi$ is assumed to be differentiable. By definition, an invariant has zero sensitivity to $G$.
 
The {\em Likelihood function} is $L(\M; \xx) \doteq p_\M(\xx)$, understood as a function of $\M$ for a fixed sample $\xx$, sometimes written as $p(\xx | \M)$ even though $\M$ is not a random variable.
Theorem 3.2 of \cite{pawitan} can be extended to an infinite-dimensional parameter $\M$ (Theorem 6.1 of \cite{bahadur1954sufficiency}): 

\begin{thm}[The likelihood function as a statistic] The likelihood function $L(\M; \xx)$ is a minimal sufficient statistic of $\xx$ for $\M$.
\label{thm-lik}
\end{thm}

\vspacetitle
\subsection{Nuisance management: Profile, marginal, and SAL likelihoods}
\label{sect-lik}
\label{sect-marg-max}
\vspacetitle

A {\em nuisance} $g\in G$ is an unknown ``factor'' (a random variable, parameter, or transformation) that is not of interest and yet it affects the data.
Given $p_\M(\cdot)$, when $g$ is treated as a parameter that transforms the data via $g(\f) \doteq g\f$, then $p_{\M,g}(\f) \doteq p_\M(g\f)$; when it is treated as a random variable, $p_{\M}(\f | g) \doteq p_\M(g\f)$. 
The {\em profile likelihood}
\be
p_{\M, G}(\f) \doteq \sup_{g\in G}p_{\M, g}(\f) %\doteq L_G(\M; \f)
\label{eq-max-out}
\ee
where the nuisance has been ``maxed-out'' is $G$-invariant. The {\em marginal likelihood} 
\be
p_{\M}(\f | G) \doteq \int_G p_\M(\f | g)dP(g) %\doteq L(\M; \f | G)
\label{marg-lik}
\ee
is invariant {\em only if} $dP(g) = d\mu(g)$ is the constant\footnote{Base, or Haar measure if $G$ is a group. It can be improper if $G$ is not compact.}  measure on $G$. Both are {\em sufficient invariants}, in the sense that they are invariant to $G$ {\em and} are minimal sufficient. This counters the common belief that ``invariance trades off selectivity.'' In Rem. \ref{rem-sufficient-invariant} we argue that both can be achieved, at the price of complexity. 

Computing the profile likelihood in practice requires reducing $G$ to a countable set $\{g_1, \dots, g_N\}$ of {\em samples},\footnote{Note that $N$ can be infinite if $G$ is not compact.} usually at a loss.  The tradeoff is a subject of sampling theory, where samples can be generated {\em regularly}, independent of the signal being sampled, or {\em adaptively}.\footnote{\label{footnoteSampling}$\{g_i\}_{i=1}^N$ are generated by a (deterministic or stochastic) mechanism $\psi$ that depends on the data and respects the structure of $G$. If $G$ is a group, this is known as a {\em co-variant detector}: It is a function $\psi$ that (is Morse in $g$, \ie it) has isolated extrema $\{g_i(\f)\}_{i=1}^N = \{g \ | \  \nabla_G \psi(\f, g) = 0\}$ that equivary: $\nabla_G\psi(\tilde g \f, g_i) = 0 \Rightarrow \nabla_G \psi(\f , \tilde g  g_i) = 0$ for all $i$ and $\tilde g \in G$. The samples $\{g_i\}_{i=1}^N$ define a reference frame in which an invariant can be easily constructed in a process known as {\em canonization} \cite{soatto09}:  $\phi(\f) \doteq \{g_i^{-1}(\f)\f \ | \ \nabla_G\psi(\f, g_i) = 0$\}.} 
In either case, the occurrence of spurious extrema (``aliases'') can be mitigated by retaining {\em not} the value of the function at the samples, $p_{\M, g_i}(\f)$, but an {\em anti-aliased} version consisting of a weighted average around the samples: 
\be
\hat p_{\M, g_i}(\f) \doteq \int p_{\M, g_i}(g\f)w(g)d\mu(g)
\ee
for suitable weights $w$.\footnote{For regular sampling of stationary signals on the real line, optimal weights for reconstruction can be derived explicitly in terms of spectral characteristics of the signal, as done in classical (Shannon) sampling theory. More in general, the computation of optimal weights for function-valued signals defined on a general group $G$ for tasks other than reconstruction, is an open problem. Recent results \cite{chen2011diffusion} show that diffusion on the location-scale group {\em typically} reduce the incidence of topological features such as aliases in the filtered signal. Thus, low-pass filtering such as (generalized) Gaussian smoothing as done here, can have anti-aliasing effects.} When the prior $dP(g) = w(g)d\mu(g)$ is positive and normalized, the previous equation (anti-aliasing) can be interpreted as {\em local marginalization}, and is often referred to as {\em mean-pooling}.  
The approximation of the profile likelihood obtained by sampling and anti-aliasing, is called the {\em SA (sampled anti-aliased) likelihood, or SAL:}
\be
\hat p_{\M, G}(\f) = \max_{i} \hat p_{\M, g_i}(\f) = \max_i \int p_{\M, g_i}(gy)dP(g).
\label{eq-soa}
\ee
The maximization over the samples in the above equation is often referred to as  {\em max-pooling}.
\begin{claim}[The SAL is an optimal representation]\label{claim-sal}
Let the joint likelihood $p_{\M, g}$ be smooth with respect to the base measure on $G$. For any approximation error $\epsilon$, there exists an integer $N = N(\epsilon)$ number of samples $\{g_i\}_{i=1}^N$ and a suitable (regular or adaptive) sampling mechanism  so that the SAL $\max_i \hat p_{\M, g_i}$ approximates to within $\epsilon$ the profile likelihood $\sup_{g \in G} p_{\M, g}$, after normalization, in the sense of distributions. 
\end{claim}
For the case of (conditionally) Gaussian models under the action of the translation group, the claim follows from classical sampling arguments. More generally, an optimal representation is difficult to compute. In the next section, we show a first example when this can be done.

\begin{rem}[Invariance, sensitivity and ``selectivity'']
\label{rem-sufficient-invariant}
It is commonly believed that invariance comes at the cost of discriminative power.  This is partly due to the use of the term invariance (or ``approximate invariance'' or ``stability'') to denote {\em sensitivity}, and of the term ``selectivity'' to denote maximal invariance. A function is {\em insensitive} to $g$ if small changes in $g$ produce small changes in its value. It is invariant to $g$ if it is constant as a function of $g$. It is a maximal invariant if equal value implies equivalence up to $G$ (Sect. 4.2 of \cite{shao98}, page 213). 
It is common to decrease sensitivity with respect to a transformation $g$ by {\em averaging} the function with respect to $g$, a lossy operation in general. For instance, if the function is the image itself, it can be made insensitive to rotation about its center by averaging rotated versions of it. The result is an image consisting of concentric circles, with much of the informative content of the original image gone, in the sense that it is not possible to reconstruct the original image. Nevertheless, while invertibility is relevant for reconstruction tasks, it is not necessarily relevant for classification, so it is possible for the averaging operation to yield a sufficient invariant, albeit not a maximal one.
\end{rem}
Thus, {\em one can have invariance while retaining sufficiency}, albeit generally not maximality: The profile likelihood, or the marginal likelihood with respect to the uniform measure, are sufficient statistics and are (strictly) invariant. The price to have both is {\em complexity}, as both are infinite-dimensional in general. However, they can be approximated, which is done by sampling in the SA likelihood.

\subsection{A first example: local representations/descriptors}

Let the task be to decide whether a (future) image $\f$ is of a scene $\M$ given a {\em single} training image $\xx$ of it, which we can then assume to be the scene itself: $\xx = \M$. Nuisances affecting $\f$ are limited to translation parallel to the image plane, scaling, and changes in pixel values that do not alter relative order.\footnote{The planar translation-scale group can be taken as a very crude approximation of the transformation induced on the image plane by spatial translation. Contrast transformations (monotonic continuous transformations of the range of the image) can be interpreted as crude approximations of changes of illumination.}  Under these (admittedly restrictive) conditions, the SAL can be easily computed and corresponds to known ``descriptors.'' Note that, by definition, $\xx$ and $\f$ must be generated by the same scene $\M$ for them to ``correspond.'' 
\label{sect-sift}

SIFT \cite{lowe04distinctive} performs canonization${}^{\ref{footnoteSampling}}$ of local similarity transformations via adaptive sampling of the planar translation-scale group (extrema of the difference-of-Gaussians operator in space and scale), and planar rotation (extrema of the magnitude of the oriented gradient). Alternatively, locations, scales and rotation can be sampled regularly, as in ``dense SIFT.'' Regardless, on the domain determined by each sample, $g_i$, SIFT computes a weighted histogram of gradient orientations. Spatial regularization anti-aliases translation; histogram regularization anti-aliases orientation; scale anti-aliasing, however, is not performed, an omission corrected in DSP-SIFT \cite{dongS15}. 

In formulas, if $\alpha(\f) = \angle \nabla \f \in {\mathbb S}^1$ is a direction, $\M = x_i$ is the image restricted to a region determined by the reference frame $g_i$, centered at $(u_i,v_i) \in \real^2$ axis-aligned and with size $s_i >0$, we have\footnote{Here $g \f(u_i,v_i) = \f(u_i+u, v_i + v) $ for the translation group and $g\f(u_i,v_i) = \f\left( \sigma u_i +u, \sigma v_i + v)\right)$ for translation-scale. In general, $g\f - \xx \neq \f - g^{-1}\xx$, unless $g\f - \xx = 0$. If $p_\xx(\f(u)) \doteq q(\f(u) - \xx(u))$ for some $q$ is a density function for the random variable $\f(u)$, in general  $q(g \f(u) - \xx(u)) \neq q(\f(u) - g^{-1}\xx(u)),$ unless the process $\f$ is $G$-stationary independent and identically distributed (IID), in which case $p_\xx(g\f) = p_{g^{-1}\xx}(\f)$. Note that the marginal density of the gradient of natural images is to a reasonable approximation invariant to the translation-scale group \cite{huangM99}.}
\be
\phi_{\xx_i}(\f)  = 
\int {\kappa}_\sigma(u_i-\tilde u, v_i -\tilde v) \kappa_\epsilon(\angle \nabla \xx(\tilde u,\tilde v), \alpha(\f)) \| \nabla \xx(\tilde u, \tilde v) \| {\cal E}_{s_i}(\sigma) d\tilde u d\tilde v d\sigma
\label{eq-dsp}
\ee
and $\phi_{\rm sift}(\alpha) = [\phi_{\xx_{11}}(\f), \dots, \phi_{\xx_{44}}(\f)]$ is a sampling on a $4\times 4$ grid, with each sample assumed independent and $\alpha = \angle \nabla \f$ quantized into $8$ levels. Variants of SIFT such as HOG differ in the number and location of the samples, the regions where the histograms are accumulated and normalized. Here $\kappa_\sigma$ and $\kappa_\epsilon$ are Parzen kernels (bilinear in SIFT; Gaussian in DSP-SIFT) with parameter $\sigma, \epsilon >0$ and ${\cal E}_s$ is an exponential prior on scales. Additional properties of SIFT and its variants are discussed in Sect. \ref{sect-contrast}, as a consequence of which the above approximates the SAL for translation-scale and contrast transformation groups.

\begin{claim}[DSP-SIFT]
\label{claim-dsp}
The continuous extension of DSP-SIFT \cite{dongS15} \eqref{eq-dsp} is an anti-aliased sample of the profile likelihood \eqref{eq-soa} for $G = SE(2)\times \real^+ \times {\cal H}$ the group of planar similarities transformations and local contrast transformations, when the underlying scene $\M = \xx_i$ has locally stationary and ergodic radiance, and the noise is assumed Gaussian IID with variance proportional to the norm of the gradient.
\end{claim}
The proof follows from a characterization of the maximal invariant to contrast transformations described in Sect.~\ref{sect-contrast}. 

Out-of-plane rotations induce a scene-shape-dependent deformation of the domain of the image that cannot be determined from a single training image, as we discuss in Sect. \ref{sect-realistic}. 

When interpreting local descriptors as samples of the SAL, they are usually assumed {\em independent}, an assumption lifted in  Sect. \ref{sect-nn} in the context of convolutional architectures.

\section{A more realistic instantiation}
\label{sect-realistic}

Relative motion between a non-planar scene and the viewer generally triggers occlusion phenomena. These call for the representation to be {\em local}. Intrinsic variability of a non-static scene must also be taken into account in the representation. In this section we describe the approximation of the SAL under more realistic assumptions than those implied by local, single-view descriptors such as SIFT.

\subsection{Occlusion, clutter and ``receptive fields''} 
\label{sect-occlusion}

 The data $\f$ has many components, $\f = \{\f_1, \dots, \f_{{}_{M_\f}}\}$, only an unknown subset of which from the scene or object of interest $\M$ (the ``visible'' set $V \subset D = \{1, \dots,M_\f \}$). The rest come from clutter, occlusion and other phenomena unrelated to $\M$, although they may be informative as ``context.'' We indicate the restriction of $\f$ to $V$ as $\f_{|_{V}} = \{\f_j, \ j \in V\}$. Since the visible set is not known, profiling  
$%\be
 p_{\M, G}(\f) = \max_{i, V \in {\cal P}(D)}  p_{\M,g_i}(\f_{|_{V}})
%= \max_{i, V \in {\cal P}(D)} \prod_{j \in V}  p_{\M, g_i}(y_j).
%\label{eq-fact}
$ %\ee
requires searching over the power set ${\cal P}(D)$.\cutForReview{One could break the combinatorial complexity by assuming that each receptive field is sufficient to locally determine a point estimate of $\M$, in the sense that $p(\f | \M) \simeq p(\f_{|_{V_j}} | \M) \ \forall \ j$, and find the visible set {\em not} by maximizing with respect to ${\cal P}(D)$ for each class $\M_k$, but instead maximizing with respect to $k$ for each receptive field, and then approximating
$V \simeq \cup_j \{ V_j \ | \ \M_k = \arg\max_\M p(\f_{|_{V_j}} | \M) \}$, ignoring shape priors.}
To make computation tractable,\cutForReview{ assuming some topology on $D$,} we can restrict the visible set $V$ to be the union of ``receptive fields'' $V_j$, that can be obtained by transforming\footnote{The action of a group $g$ on a set ${\cal B} \subset D$ is defined as $g{\cal B} \subset D$ such that $g(\f_{|_{\cal B}}) = \f_{|_{g {\cal B}}}$.} a ``base region''  ${\cal B}_0$ (``unit ball,'' for instance a square patch of pixels with an arbitrary ``base size,'' say $10\times 10$) with group elements $g_j \in G$, $V_j \doteq g_j {\cal B}_0$:
$% \be
V = \bigcup_{j = 1}^M g_j {\cal B}_0
%\label{eq-receptive}
$ % \ee
where the number of receptive fields $M \ll M_\f$\cutForReview{ is much smaller than the number of components of $\f$}. Thus $V$ is determined by the reference frames (group elements) $\{g_j\}_{j = 1}^M$ of receptive fields that are ``active,'' which are unknown a-priori.\cutForReview{ With an abuse of notation, we refer to $V$ as the set of indices $j$ of the active receptive fields, rather than the index of every pixel in every active receptive field.}

Alternatively, we can marginalize $V$ by computing, for every class (represented by a hidden variable $\M_k$ as discussed next) and every receptive field (determined by $g_j$ as above), conditional densities $ p_\M(\f | \M_k, g_j)$ that can be averaged with respect to weights $w_{jk}$, trained to {\em select} (if sparse along $j$) and {\em combine} (by averaging along $k$) local templates via
\be
\sum_{j,k}  p_\M(\f | \M_k, g_j) w_{jk} 
\ee
where the weights or ``filters'' $w_{jk}$, if positive and normalized,\footnote{Note that current convolutional architectures rectify and normalize the feature maps, not the filters. However, learned biases, as well as rectification and normalization at each layer, may partly compensate for it.} are interpreted as probabilities $w_{jk} = p_\M(\M_k, g_j )$. To make this marginalization tractable, we write the first term as 
\be\label{eq:nocontext}
 p_\M(\f_{|_{V_j}},\f_{|_{V^c_j}} | \M_k, g_j) = p(\f_{|_{V_j}} | \M_k, g_j)  p_\M(\f_{|_{V^c_j}}) \propto  p(\f_{|_{V_j}} | \M_k, g_j)\ee where we have assumed that the second factor is constant, thus ignoring {\em photometric context} beyond the receptive fields, \eg due to mutual illumination.\cutForReview{ This approximation may be responsible for the effects shown in Fig.~1 of \cite{karianakisDS15}.} Under these assumptions, we have
\be
  p_{\M,g_i}(\f) = \sum_{j,k} p(\f_{|_{V_j}} | \M_k, g_i g_j)  p_{\M,g_i}(g_j \M_k) %= \sum_{j,k} p_{\M_k, g_{ij}}(\f_{|_{V_j}})  p_{\M, g_{ij}}(\M_k)
\label{eq-receptive}
\ee
where the first term in the sum is known as a {\em ``feature map''} and we have assumed that both $g_i, g_j \in G$ for simplicity, with $g_{ij} = g_i g_j$. \cutForReview{ In particular, since we assume that nuisance variability affecting the {\em range space} of the images is restricted to contrast transformations, addressed in Sect. \ref{sect-contrast}, the rest is restricted to domain transformations that, once restricted to bounded domains, can be approximated by low-dimensional groups.}
The order of  operations (deformation by $g_i$ and selection by $g_j$) is arbitrary, so we \cutForReview{must choose a convention. We} assume that the selection by $g_j$ is applied first, and then the nuisance-induced deformation $g_i$, so $ p_{\M, g_i}(g_j \f) \propto  p_{\M,e}(g_{ij}\f)$, where $e$ is the identity of the group $G$.

\subsection{Intra-class variability and deformable templates} 
\label{sect-deformable}

For category-level recognition, the parameter space can be divided into $K$ classes, allowing variability of $\M$ within each class. Endowing the parameter space with a distribution $p(\M | k)$ \cutForReview{is laborious. For the case of object detection in 3-D, it} requires defining a probability density in the space of shapes, reflectance functions etc. \cutForReview{If the class $k$ can be represented by an orbit under a group $G$, one can exploit a density on the group to define a class-conditional density on $\M$. To this end, pick any representative of the class, say $\M_k \sim p(\M | k)$, then marginalize with respect to the density $ p(g | \M_k)$, to obtain\footnote{The last equality follows from the assumption that $G$ acts transitively on $\M_k$, and thus $p(\M | g\M_k) = \delta (\M - g\M_k)$.}
$
p(\M | k) = \int p(\M | \M_k, g) dP(g | \M_k) = p(g  | \M = g\M_k).
$
This is still complex when the parameter space $\M$ is non-trivial.} Alternatively one can capture the variability $\M$ induces on the data. For any scene $\M_k$ from class $k$, one can consider a single image generated by it $\xx_k \sim p_{\M_k}(\xx)$ as a ``template'' from which any other datum from the same class can be obtained by the (transitive) action of a group  $g_k\in G$.\cutForReview{ Note that now the group is acting on the data. In general even a simple group acting on the scene (\eg $SE(3)$) induces a very complex action on the data (\eg diffeomorphisms \cite{sundaramoorthiPVS09}). Nevertheless, we use the same symbol $G$ and will show later that this entails no loss of generality thanks to the restriction to receptive fields.} Thus if $\f \sim p_{\M_k}(\f)$ with $\M_k \sim p(\M | k)$, which we indicate with $\f \sim p(\f | k)$, then we assume that there exists a $g_k$ such that $\f = g_k^{-1}\xx_k$, so 
\bea
p(\f | k) &=& \int p(\f | \xx_k, g_k) dP(\xx_k, g_k | \M_k) dP(\M_k | k) \nonumber \\
%&=& \int \delta(\f - g_k^{-1}\xx_k) dP(\xx_k | \M_k')dP(g_k | \M_k') \delta(\M_k'-\M_k) \\
&=& \int p(g_k \f | \M_k)dP(g_k | \M_k)  %\\
%&=& \int dP_{\xx_k}(g_k \f | \M_k )dP(g_k | \M_k) %= \int p_{\M_k ,g_k}(\f) dP_{\M_k}(g_k)
\label{eq-template}
\eea 

where we have used the fact that $p(\f | \xx_k, g_k, k) = \delta(\f - g^{-1}_k x_k)$ and that only one sample of $\M_k$ is given, so all variability is represented by $g_k$ and $x_k$ conditionally on $\theta_k$.\footnote{In the last expression we assumed that $x_k$ and $g_k$ are conditionally independent given $\theta_k$, \ie that the image formation process (noise) and deformation are independent once the scene $\theta_k$ is given.} For this approach to work, $g_k$ has to be sufficiently complex to allow $\xx_k$ to ``reach'' every datum $\f$ generated by an element\footnote{The group has to act transitively on $\xx_k$. For instance, in \cite{grenander93} $g_k$ was chosen to belong to the entire (infinite-dimensional) group of domain diffeomorphisms.} of the class. Fortunately,\cutForReview{ because the domain has been broken into a number $M$ of receptive fields $g_j$, the action of a complex group restricted to them can be approximated by a simple group, for instance the affine or similarity group.  \cutForReview{In particular, the receptive fields can always be chosen sufficiently small that the groups acting on the domain and range can be approximated arbitrarily well by the affine or similarity group $G$.} Then} the density on a complex group can be reduced to a joint density on $G^M$, the mutual configuration of the receptive fields, as we will show. 

The restriction of\cutForReview{ the group action} $g_k$ to the domain of the receptive field $V_j = g_j{\cal B}_0$ is indicated by $\{g_{kj}\}_{j=1}^M$, defined by $g_{k_j} \xx = g_k \xx \ \forall \ \xx \in V_j$. Then, we can consider the global group nuisance $g_i$, the selector of receptive fields $g_j $ and the local restriction of the intra-class group $g_k$, assumed ($d(g_k,e)$) small, as all belonging to the same group $G$, for instance affine or similarity transformations of the domain and range space of the data. 

Starting from \eqref{eq-receptive}, neglecting $g_i$ for the moment, we have\footnote{Here we condition on the restrictions $g_{k_j}$ of $g_k$ on the receptive fields $V_j$ so that, by definition, $p(\f_{|_{V_j}} | \M_k, g_j, g_{k}) = p(\f_{|_{V_j}} | \M_k, g_{k_j})$.}
\bea
p_\M(\f) &=& \sum_{j,k} p(\f_{|_{V_j}} | \M_k, g_j)p_\M(\M_k, g_j) \\
%&=&  \sum_{j,k} \int_{{\rm diff}(D)} p(\f_{|_{V_j}} | \M_k, g_j, g_k)dP(g_k | \M_k, g_j)p(\M_k, g_j | \M) \\
%&=&  \sum_{j,k,l} \int p(\f_{|_{V_j}} | \M_k, g_j, g_k, \{g_{k_l}\})dP(g_k, \{g_{k_l}\} | \M_k, g_j)p(\M_k, g_j | \M) \\
%&=&  \sum_{j,k,l} \int p(\f_{|_{V_j}} | \M_k, g_j, g_k, \{g_{k_l}\})\prod_l \delta(g_k g_j g_{k_l}^{-1})dP(\{g_{k_l}\} | \M_k)p(\M_k, g_j | \M) \\
&=&  \sum_{j,k} \int_{G^M} p(\f_{|_{V_j}} | \M_k, g_{k_j})dP(\{g_{k_j}\} | \M_k)p_\M(\M_k, g_j)
\label{eq-almost-all}
\eea
and bringing back the global nuisance $g_i$, 
\be
 p_{\M, G}(\f) =  \max_i  \underbrace{\sum_{j,k} \int_{G^M} p (g_{ik_j} \f_{|_{V_j}} | \M_k)dP_G(\{g_{k_j}\} | \M_k)p_{\M}(\M_k, g_j)}_{  p_{\M}(g_i \f)}
\label{eq-all}
\ee
where the measure in the last equation is made invariant to $g_i \in G$. The feature maps $p (g_i g_{k_j} \f_{|_{V_j}} | \M_k)$ represent the photometric component of the likelihood.
The geometric component\cutForReview{ of the likelihood} is the relative configuration of receptive fields $\{g_{k_j}\}$, which is class-dependent but $G$-invariant. The inner integral corresponds to ``mean pooling'' and the maximization to ``max pooling.'' The sum over $j,k$ marginalizes the local classes $\M_k$, or {\em ``parts''} and selects them to compose the hypothesis $\M$.

To summarize, $g_i$ are the samples of the nuisance group in \eqref{eq-soa}; $g_j$ are the local reference frames that define each receptive field in \eqref{eq-receptive}; $g_k$ are the global deformations that define the variability induced by a class $k$ on a template in \eqref{eq-template}. The latter are in general far more complex than the former, but their restriction to each receptive field, $g_{k_j}$, can be approximated by an affine or similarity transformation and hence composed with $g_i$ and $g_j$.

Note that   \eqref{eq-almost-all} can be interpreted as a model of a three-layer neural network: The visible layer, where $\f$ lives, a hidden layer, where the feature maps $p(\f_{|_{V_j}} | \M_k, g_{k_j})$ live, and an output layer that, after rectification and normalization, yields an approximation of the likelihood $p_\M(\f)$.  Invariance to $G$ can be obtained via  a fourth layer outputting $ p_{\M, G}(\f)$ by max-pooling third-layer outputs $p_{\M}(g_i \f)$  for different $g_i$ in \eqref{eq-all}.

\cutForReview{\vspacetitle
\subsection{Learning Visual Representations} 
\vspacetitle}
\label{sect-nuisances}

\vspacetitle
\subsection{Contrast invariance} 
\label{sect-contrast}
\vspacetitle

\begin{figure}[htb]
\begin{center}
\includegraphics[height=.21\textwidth,width=.315\textwidth]{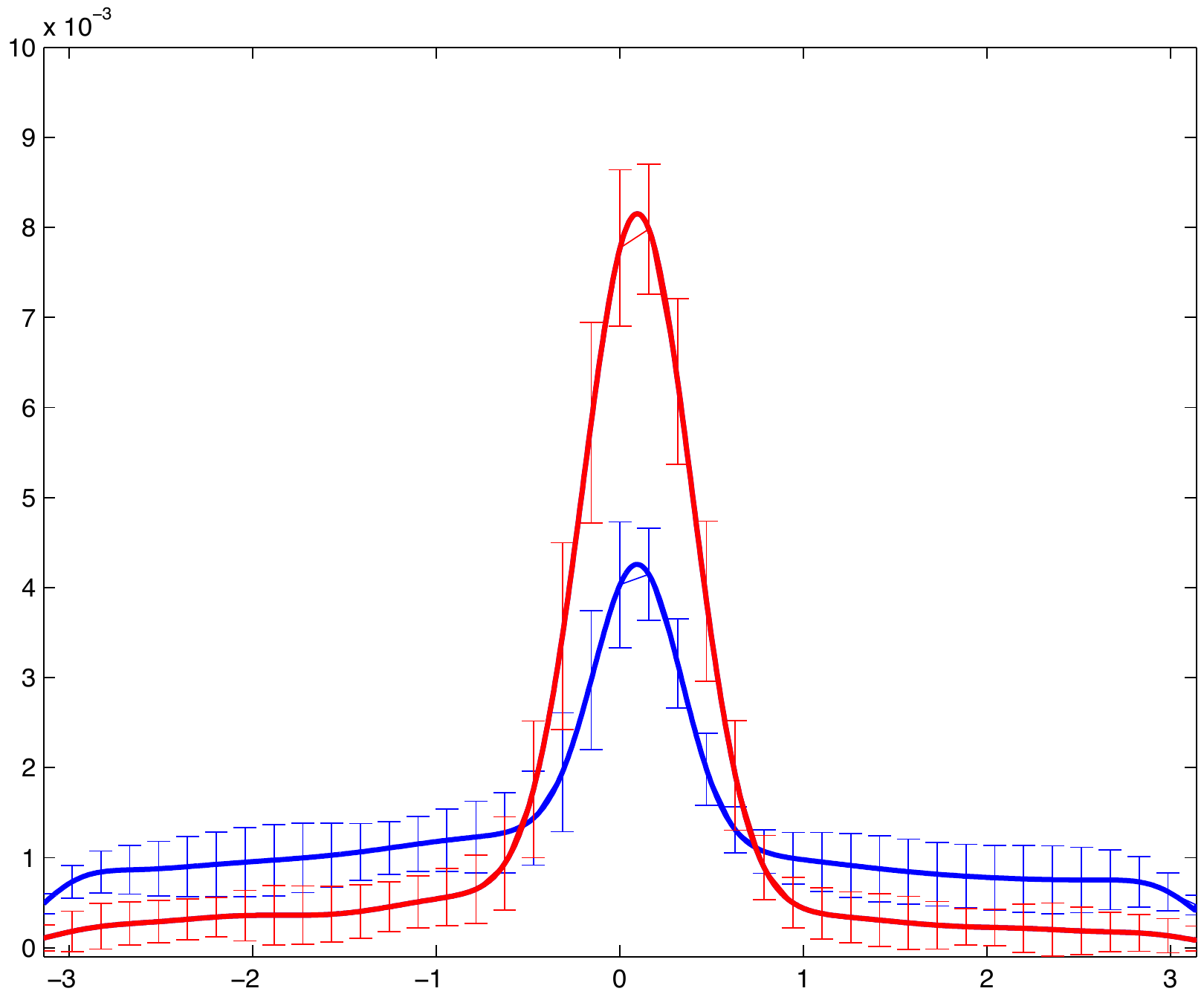}
\includegraphics[height=.21\textwidth,width=.315\textwidth]{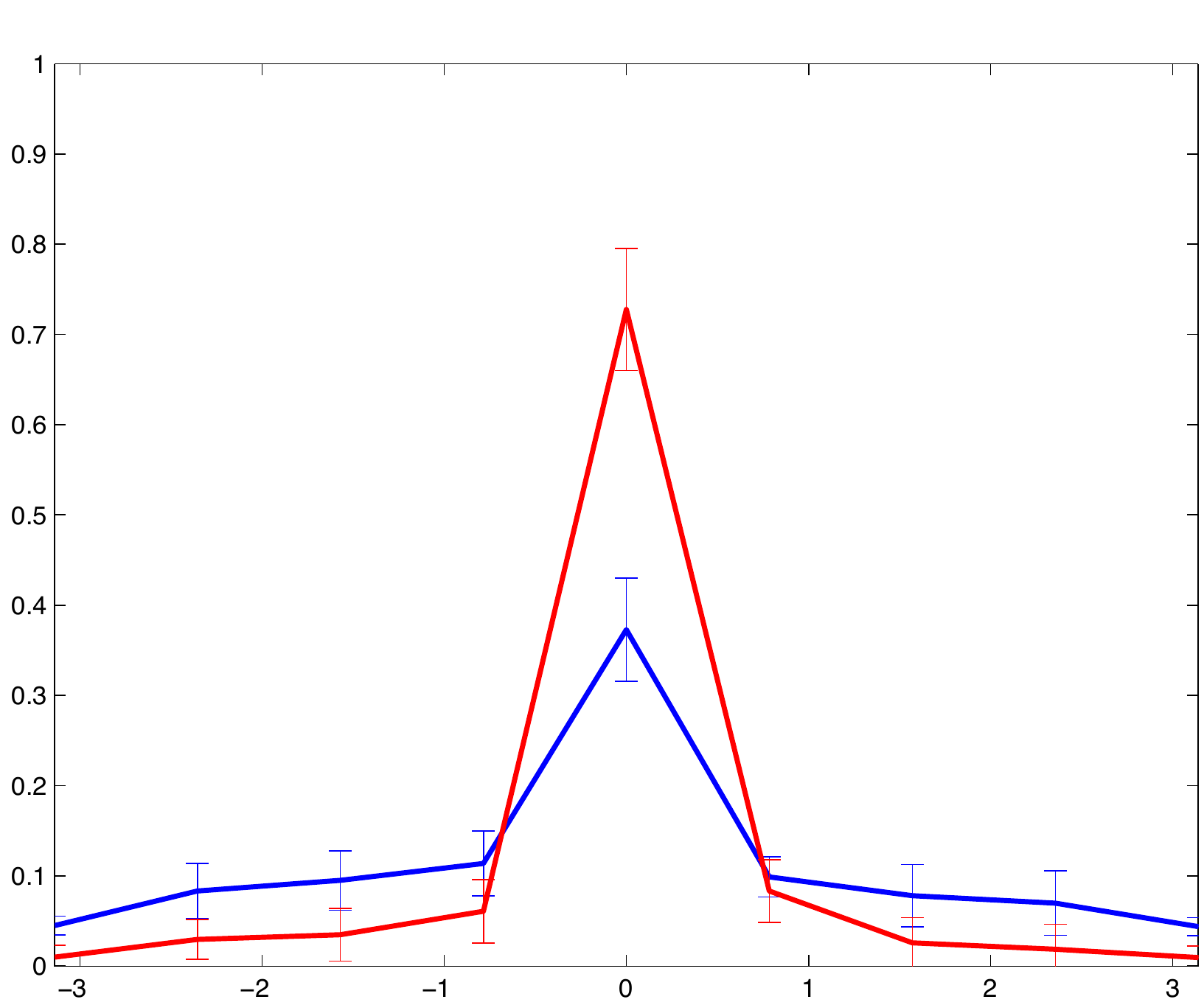}
\end{center}
\vspace{-.5cm}
\caption{\sl {\bf SIFT integrand \eqref{eq-sift-integrand} (red) vs. marginalized likelihood \eqref{eq-contr-inv} (blue)} computed for a random patch on $\alpha \in [-\pi, \pi]$ (left), and on a regular sub-sampling of $8$ orientations (right). Several random tests are shown as mean and error-bars corresponding to three standard deviations across trials.}
\label{fig-sift-no-good}
\end{figure}

Contrast is a monotonic continuous transformation of the (range space of the) data, which can be used to model changes due to illumination. It is well-known that the curvature of the level sets of the image \cutForReview{ at each point} is a maximal invariant \cite{alvarezGLM93}. Since it is everywhere orthogonal to the level sets, the gradient orientation is also a maximal contrast invariant. Here we compute a contrast invariant by marginalizing the norm of the gradient of the test image (thus retaining its orientation) in the likelihood function of a training image.  Since the action of contrast transformations is spatially independent, in the absence of other nuisances we assume that the gradient of the test image $\f$ can be thought of as a noisy version of the gradient of the training image $\xx$,  \ie 
\be
\label{eq-mod-ass}
\nabla \f \sim  {\cal N}(\nabla \xx,\epsilon^2)
\ee
and compute the density of $\f$ given $\xx$ marginalized with respect to contrast transformations ${\cal H}$ of $\f$. 
\begin{thm}[Contrast-invariant sufficient statistic]
\label{claim-contrast}
The likelihood of a training image $\xx$ at a given pixel, given a test image $\f$, marginalized with respect to contrast transformations of the latter, is given by 
\be \label{eq-contr-inv}
\boxed{p_{\xx}(\f | {\cal H}) \doteq p(\angle \nabla \f | \nabla \xx) = \frac{1}{\sqrt{2\pi\epsilon^2}} \exp\left(
-\frac{1}{2\epsilon^2}\sin^2(\angle \nabla \f - \angle \nabla \xx) \|\nabla \xx \|^2
\right) M}
\ee
\def\m{\cos(\angle \nabla \f - \angle \nabla \xx) \| \nabla \xx \| }
where, if we call $\Psi(a) \doteq \frac{1}{\sqrt{2\pi}} \int_{-\infty}^{a}  e^{-\frac{1}{2}\tau^2} \,d\tau$  for any $a \in \real$, and $m \doteq \m$, then 
\def\m{m}
\be \label{eq-M}
%M  = \frac{\epsilon e^{-\frac{ (\m)^2 }{2\epsilon^2}}}{\sqrt{2\pi}} + \m \left(1-\Psi\left(-\frac{\m}{\epsilon}\right)\right).
M  = \frac{\epsilon e^{-\frac{ (\m)^2 }{2\epsilon^2}}}{\sqrt{2\pi}} + \m  - \m \Psi\left(-\frac{\m}{\epsilon}\right).
\ee
The expression in \eqref{eq-contr-inv} is, therefore, a minimal sufficient statistic of $\f$ that is invariant to contrast transformations.  
\end{thm}
\cutForReview{
Comparison of \eqref{eq-contr-inv} with SIFT \cite{lowe04distinctive} is described in Remark \ref{rem-sift}, but already at the outset we notice that the latter is neither a density (it does not integrate to one as the angle $\angle \nabla \f$ spans the circle), nor does it reduce to uniform when the training patch is flat ($\| \nabla \xx \| = 0$). Fig. \ref{fig-sift-no-good} compares the two as functions of $\ttheta = \angle \nabla \f$ for randomly sampled patches $\xx$ in natural images.} 

\begin{rem}[Relation to SIFT]
\label{rem-sift}
Compared to \eqref{eq-contr-inv}, SIFT (i) neglects the normalization factor $\frac{M}{\sqrt{2\pi}\epsilon}$, (ii) replaces the kernel 
\be
\tilde \kappa_\epsilon(\alpha) \doteq \exp\left(\frac{1}{2\epsilon^2}\sin^2 (\alpha)\right)
\simeq \exp\left(\frac{1}{2\epsilon^2}\alpha^2\right)
\label{eq-tilde-kappa}
\ee
with a {\em bilinear} one $\kappa_\epsilon$ defined by 
\be
\kappa_\epsilon(\alpha) \doteq 
\begin{cases}
\frac{\alpha + \epsilon}{\epsilon^2} \quad \alpha \in [-\epsilon, \ 0] \\
\frac{\epsilon - \alpha}{\epsilon^2} \quad \alpha \in [0, \ -\epsilon] \\
\end{cases}
\label{eq-kappa}
\ee 
and, finally, (iii) multiplies the result by the norm of the gradient, obtaining the {\em sift integrand} 
\be
\phi_{\rm sift}(\angle \nabla \f | \nabla \xx) = \kappa_\epsilon(\angle \nabla \f - \angle \nabla \xx) \| \nabla \xx \|
\label{eq-sift-integrand}
\ee
To see this, calling $\alpha = \angle \nabla \f - \angle \nabla \xx$ and $\beta = \| \nabla \xx \| > 0$, notice that %$\kappa_\epsilon (\alpha \beta) =  \kappa_{\frac{\epsilon}{\beta}} (\alpha) \beta$, so
\be
\kappa_\epsilon(\alpha) \beta  = \kappa_{\epsilon \beta }(\alpha \beta ) \simeq 
\exp \left( \frac{1}{2\epsilon^2 \beta^2} \alpha^2 \beta^2 \right) \simeq \tilde \kappa_\epsilon(\alpha)
\ee
where the left-hand side is \eqref{eq-sift-integrand} and the right-hand side is \eqref{eq-contr-inv} or \eqref{eq-quasi-sift}.

We make no claim that this approximation is good, it just happens to be the choice made by SIFT, and the above just highlights the relation to the contrast invariant \eqref{eq-contr-inv}.
\end{rem}

\begin{rem}[Uninformative training images]
\cutForReview{One may observe that the bracketed expression admits simple approximations, \eg 
$$
M = \int_0^{\infty} \frac{1}{\sqrt{2\pi\epsilon^2}} e^{-\frac{1}{2\epsilon^2}\left(\rho - \gamma \langle\overline{\nabla\f},\overline{\nabla\xx}\rangle\right)^2}\, \rho d\rho \simeq \left\{\begin{array}{cc} 
m & \frac{m}{\epsilon} \gg0 \\  \frac{\epsilon}{\sqrt{2\pi}} &  \frac{m}{\epsilon} \simeq 0 \\ 0  &  \frac{m}{\epsilon} \ll 0.
\end{array}\right.
$$
Note, however, that $p(\ttheta|\nabla\xx)$ is to be computed for all $\ttheta \in [0,2\pi]$; as such, conditionally on $\nabla\xx$, the quantity $\langle\overline{\nabla\f},\overline{\nabla\xx}\rangle$ ranges on the whole interval $[-1,1]$. Therefore, 
$\frac{m}{\epsilon}$ can never be uniformly $\gg 1$ or $\ll 1$.}  It is possible that $\frac{m}{\epsilon}\simeq 0$  holds uniformly (in $\ttheta$) provided  $\frac{\gamma}{\epsilon} \ll 1$, \ie,  if the modulus of the gradient $\nabla \xx$ is very small as compared to the standard deviation $\epsilon$. Under such circumstances, the training image $\xx$ is essentially constant (``flat''), and the conditional density $p(\ttheta|\nabla\xx)$ becomes uniform
\be\label{eq-uniform-alpha}
\begin{array}{rcl}
p(\ttheta|\nabla\xx)& \simeq & \frac{1}{\sqrt{2\pi\epsilon^2}} e^{-\frac{1}{2\epsilon^2}\gamma^2\left(1-\langle\overline{\nabla\f},\overline{\nabla\xx}\rangle^2\right) }  \frac{\epsilon}{\sqrt{2\pi}} \\
& = & \frac{1}{\sqrt{2\pi\epsilon^2}} \frac{\epsilon}{\sqrt{2\pi}}  = \frac{1}{2\pi}
\end{array}
\ee
where the approximation holds given that $e^{-\frac{1}{2\epsilon^2}\gamma^2\left(1-\langle\overline{\nabla\f},\overline{\nabla\xx}\rangle^2\right) }\simeq 1$ when $\frac{\gamma}{\epsilon} \ll 1$. This is unlike SIFT \eqref{eq-sift-integrand}, that becomes zero when the norm of the gradient goes to zero.
\end{rem}
Note that, other than for the gradient, the computations in \eqref{eq-contr-inv} can be performed point-wise, so for an image or patch with pixel values $\f_i$, if $\ttheta_i(\f) \doteq \angle \nabla \f_i$, we can write 
\be
p(\ttheta | \nabla \xx) 
= \prod_i p(\ttheta_i | \nabla \xx_i).
\label{eq-contr-inv2}
\ee
We often omit reference to contrast transformations $\cal H$ in $p_\xx(\f | {\cal H})$, when the argument $\ttheta$ makes it clear we are referring to a contrast invariant.  The width of the kernel $\epsilon$ is a design (regularization) parameter. 

\begin{rem}[Invariance for $\xx$]\label{invx}
Note that \eqref{eq-contr-inv2} is invariant to contrast transformations of $\f$, but {\em not} of $\xx$. This should not be surprising, since high-contrast training patches should yield tests with high confidence, unlike low-contrast patches. However, when the training set contains instances that are subject to contrast changes, such variability must be managed. 

To eliminate the dependency on $\| \nabla \xx\|$, consider a model where the noise is proportional to the norm of the gradient:
\be
\nabla \f \sim {\cal N} \left(\nabla \xx,  \tilde \epsilon^2\right)
\ee
where $\tilde \epsilon(\| \nabla \xx \|) = \epsilon \| \nabla \xx \|$. Under this noise model, the sufficient contrast invariant \eqref{eq-contr-inv} becomes 
\be
{p_{\xx}(\f | {\cal H}) \doteq p(\angle \nabla \f | \nabla \xx) = \frac{1}{\sqrt{2\pi \epsilon^2}} \exp\left(
-\frac{1}{2 \epsilon^2}\sin^2(\angle \nabla \f - \angle \nabla \xx) 
\right) M}
\label{eq-quasi-sift}
\ee
and $M$ has the same expression \eqref{eq-M} but with $m = \cos(\angle \nabla \f - \angle \nabla \xx)$. Thus a simple approach to managing contrast variability of $\xx$ {\em in addition to} $\f$ is to use the above expression in lieu of \eqref{eq-contr-inv}. 

\begin{rem}[Joint normalization]\label{rem-joint}
If we consider only {\em affine} contrast transformations $ax+b$ where $a,b$ are  assumed to be  constant on a patch $V$ which contains all the cells $C_i$ where the descriptors are computed\footnote{SIFT divides each patch into a $4\times 4$ grid of cells $C_i$, $i=1,..,16$}  it is clear that to recapture invariance w.r.t. the scale factor $a$ it is necessary and sufficient that $p(\angle \nabla \f | \nabla \xx(v_i)) = p(\angle \nabla \f | a\nabla \xx(v_i))$, $\forall v_i \in V$. 
We shall now illustrate how this invariance can be achieved.

Assume  that
data generating model \eqref{eq-mod-ass} is replaced by the distribution-dependent  model 
\be
\label{eq-mod-ass-meangrad}
\nabla \f \sim  {\cal N}(\nabla \xx,\epsilon^2(p_x)) \quad \epsilon^2(p_x) = \sigma^2 \E_{x}\| \nabla\xx\|^2 =  \sigma^2 \int \|\nabla \xx\|^2p_x(\nabla \xx)d\nabla\xx
\ee
where the noise variance $\epsilon^2$ depends linearly on  the average squared gradient norm  (w.r.t. the distribution $p_x(\nabla \xx)$);  $\sigma^2$ is   fixed constant.
The resulting marginal distribution for the gradient orientation becomes
\begin{equation}\label{eq-contr-inv-meangrad}
\bar p_{\xx}(\f | {\cal H}) \doteq \bar p(\angle \nabla \f | \nabla \xx) = \frac{1}{\sqrt{2\pi\sigma^2}} \exp\left(
-\frac{1}{2\sigma^2}\sin^2(\angle \nabla \f - \angle \nabla \xx)\frac{ \|\nabla \xx \|^2}{ \E_{x}\| \nabla\xx\|^2}
\right) \bar M
\ee
\def\barm{\cos(\angle \nabla \f - \angle \nabla \xx) \frac{\| \nabla \xx \| }{\sqrt{ \E_{x}\| \nabla\xx\|^2}}}
where, defining $\bar m \doteq \barm$,  
\def\m{\bar m}
\be %\label{eq-M}
%M  = \frac{\epsilon e^{-\frac{ (\m)^2 }{2\epsilon^2}}}{\sqrt{2\pi}} + \m \left(1-\Psi\left(-\frac{\m}{\epsilon}\right)\right).
\bar M  = \frac{\sigma e^{-\frac{ (\bar m)^2 }{2\sigma^2}}}{\sqrt{2\pi}} + \bar m  - \bar m \Psi\left(-\frac{\bar m}{\sigma}\right).
\ee
\end{rem}
Equation \eqref{eq-contr-inv-meangrad}  is clearly invariant to affine transformations of the image values $ x(v) \rightarrow a x(v) + b$, $\forall v\in V$.\footnote{Note that an
 affine transformation on the image values $ x(v) \rightarrow a x(v) + b$, $\forall v\in V$, induces a scale transformation on the distribution $p_x(\nabla\xx)$ so that 
$p_{ax+b}(\rho) = \frac{1}{a}p_x(\rho/a)$ and therefore
the average squared gradient is scaled by $a^2$, i.e. $\E_{ax+b}\|\rho\|^2 = a^2 \E_x \|\rho\|^2$.}
  It is a trivial calculation to show that using $\bar p(\angle \nabla \f | \rho) $ in lieu of $ p(\angle \nabla \f | \rho) $, the result is invariant w.r.t affine transformations.\cutForReview{In fact:
\begin{equation}\label{marglikINV}
\displaystyle{\begin{array}{rcl}
\bar p_C(y)& :=&\int \bar p(\angle \nabla \f | \rho) p_x(\rho) \, d\rho  \\
& = &  \int \bar p(\angle \nabla \f | a\rho )   p_{x}(\rho) \, d\rho\\
& = &  \int \bar p(\angle \nabla \f | \gamma )   p_{x}(\gamma/a) \, \frac{1}{a} d\gamma \quad \quad \gamma:=a \rho \\
& =& \int  \bar p(\angle \nabla \f | \gamma)   p_{ax+b}(\gamma) \, d\gamma 
\end{array}}
\ee
where the second equality stems from the fact that $p(\angle \nabla \f | a\rho )=p(\angle \nabla \f | \rho)$ holds for any $a\neq 0$.}

To obtain a sampled version of this normalization  the expected squared gradient norm can be replaced with  the sample average on the training patch 
$$
\hat \rho^2 \doteq \frac{1}{N_{pix}} \sum_{i=1}^{N_{pix}} \|\nabla \xx(v_i)\|^2
$$
so that \eqref{eq-mod-ass-meangrad} becomes, 
\be
\label{eq-mod-ass-saplemeangrad}
\nabla \f \sim  {\cal N}(\nabla \xx,\epsilon^2(V)) \quad \epsilon^2(V) = \sigma^2 \hat \rho^2 =  \sigma^2  \frac{1}{N_{pix}} \sum_{i=1}^{N_{pix} }\|\nabla \xx(v_i)\|^2
\ee
where  $v_i \in V$, $i=1,..,N_{pix}$ are the pixel locations in the training patch $V$. This procedure is known as {\em ``joint normalization''}, and is simply equivalent to normalizing the patch in pre-processing by dividing by the average gradient norm. 
\end{rem}

\subsubsection{Clamping Gradient Orientation Histograms} 
\label{sect-clamping}

Local descriptors such as SIFT commonly apply a {\em ``clamping''} procedure to modify a (discretized, spatially-pooled, un-normalized) density $\phi_{\rm sift}$ of the form \eqref{eq-sift-integrand}, by clipping it at a certain threshold $\tau$, and re-normalizing:
\be
\phi_{\rm clamp}(\ttheta | \xx) = \frac{\min \{ \phi_{\rm sift}(\ttheta | \xx), \tau \}}{\int_{\mathbb S^1} \min \{ \phi_{\rm sift}(\ttheta | \xx), \tau \}d\alpha}
\label{eq-clamp}
\ee
where $\ttheta = \angle \nabla \f \in {\mathbb S}^1$ is typically discretized into $8$ bins and $\tau$ is chosen as a percentage of the maximum, for instance  $\tau = 0.2 * \max_\ttheta \phi_{\rm sift}(\ttheta | \xx)$. Although clamping has a dramatic effect on performance, with high sensitivity to the choice of threshold, it is seldom explained, other than as a procedure to ``reduce the influence of large gradient magnitudes'' \cite{lowe04distinctive}.

\begin{figure}[htb]
\begin{center}
\includegraphics[height=.1\textwidth,width=.15\textwidth]{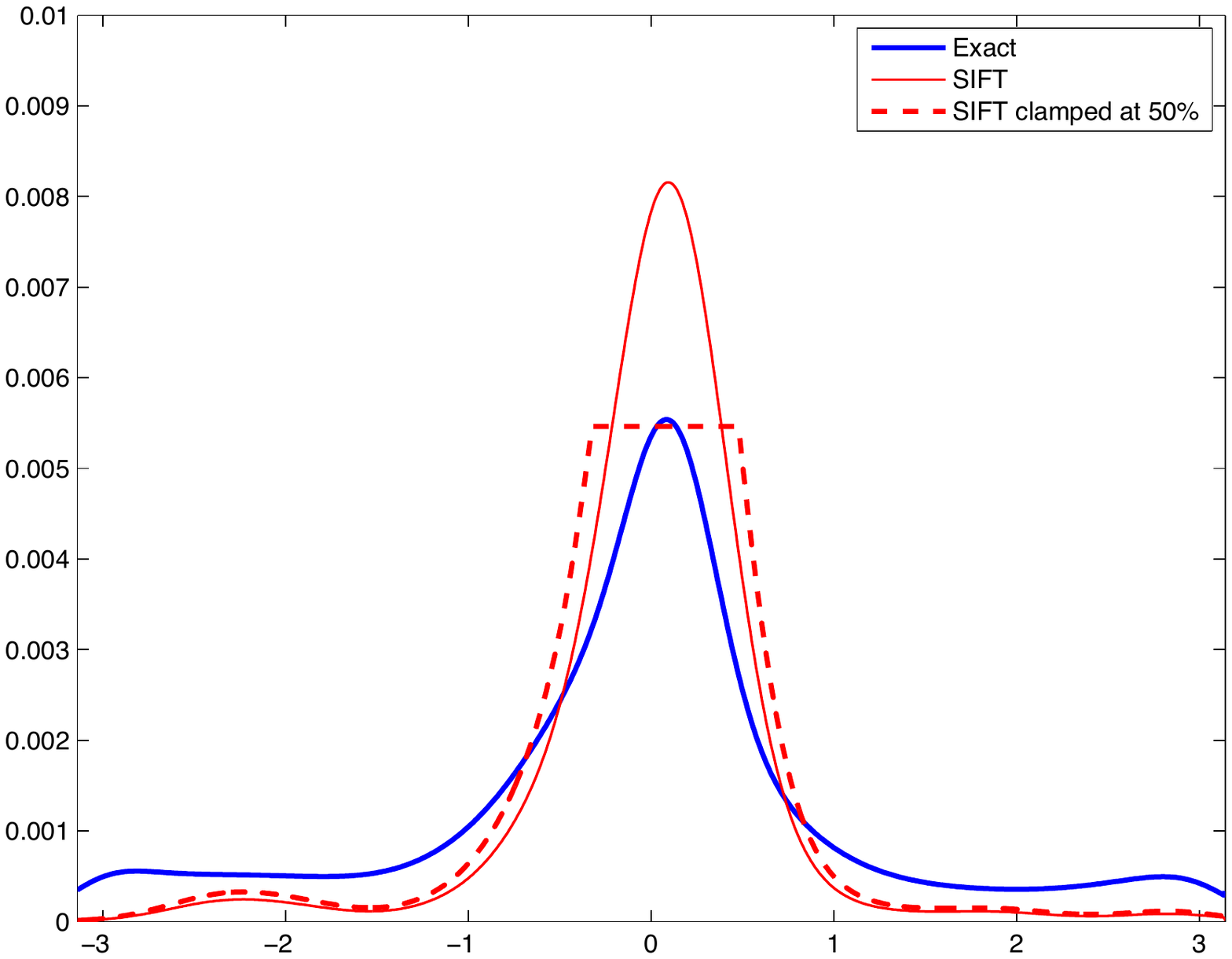}
\includegraphics[height=.1\textwidth,width=.15\textwidth]{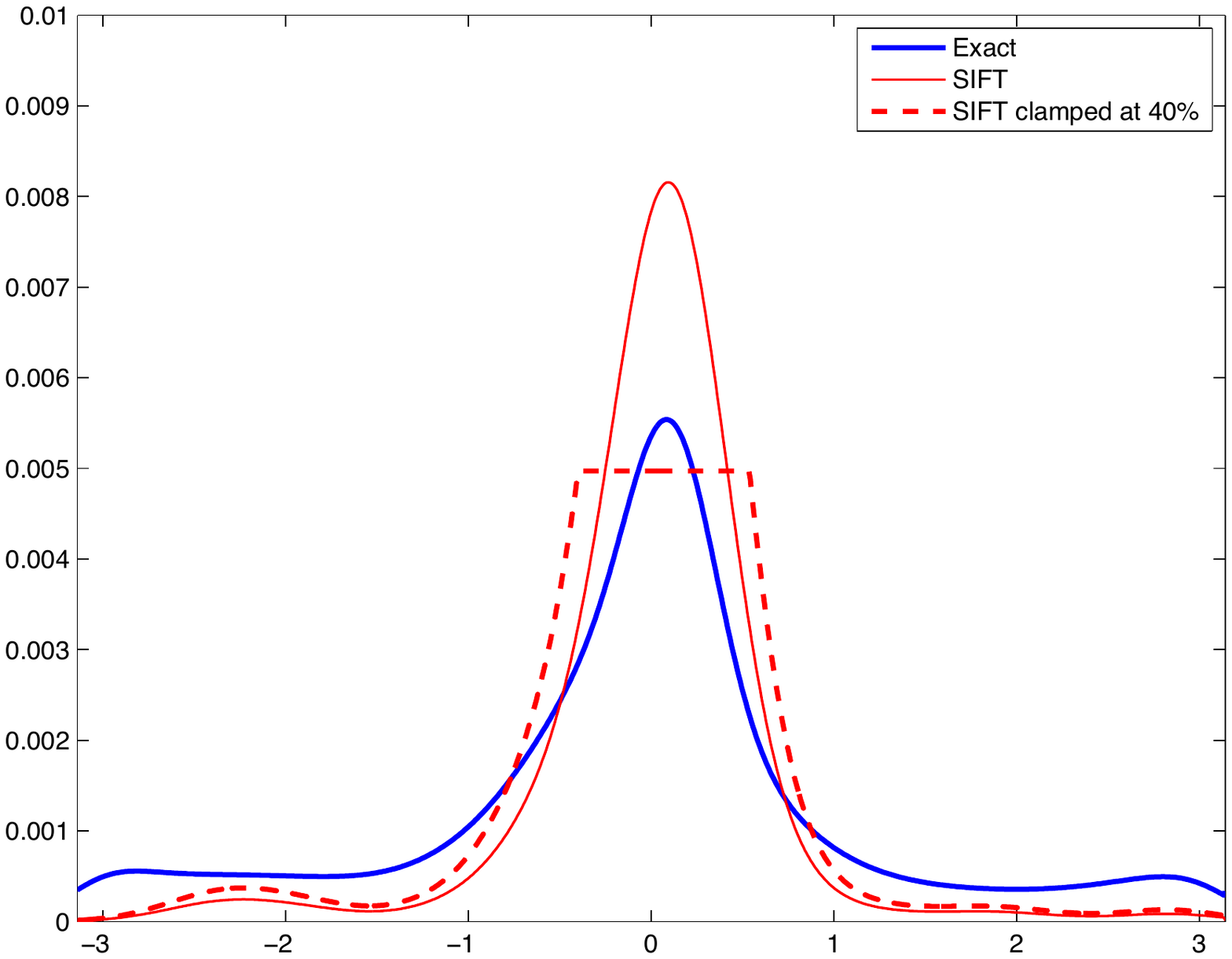}
\includegraphics[height=.1\textwidth,width=.15\textwidth]{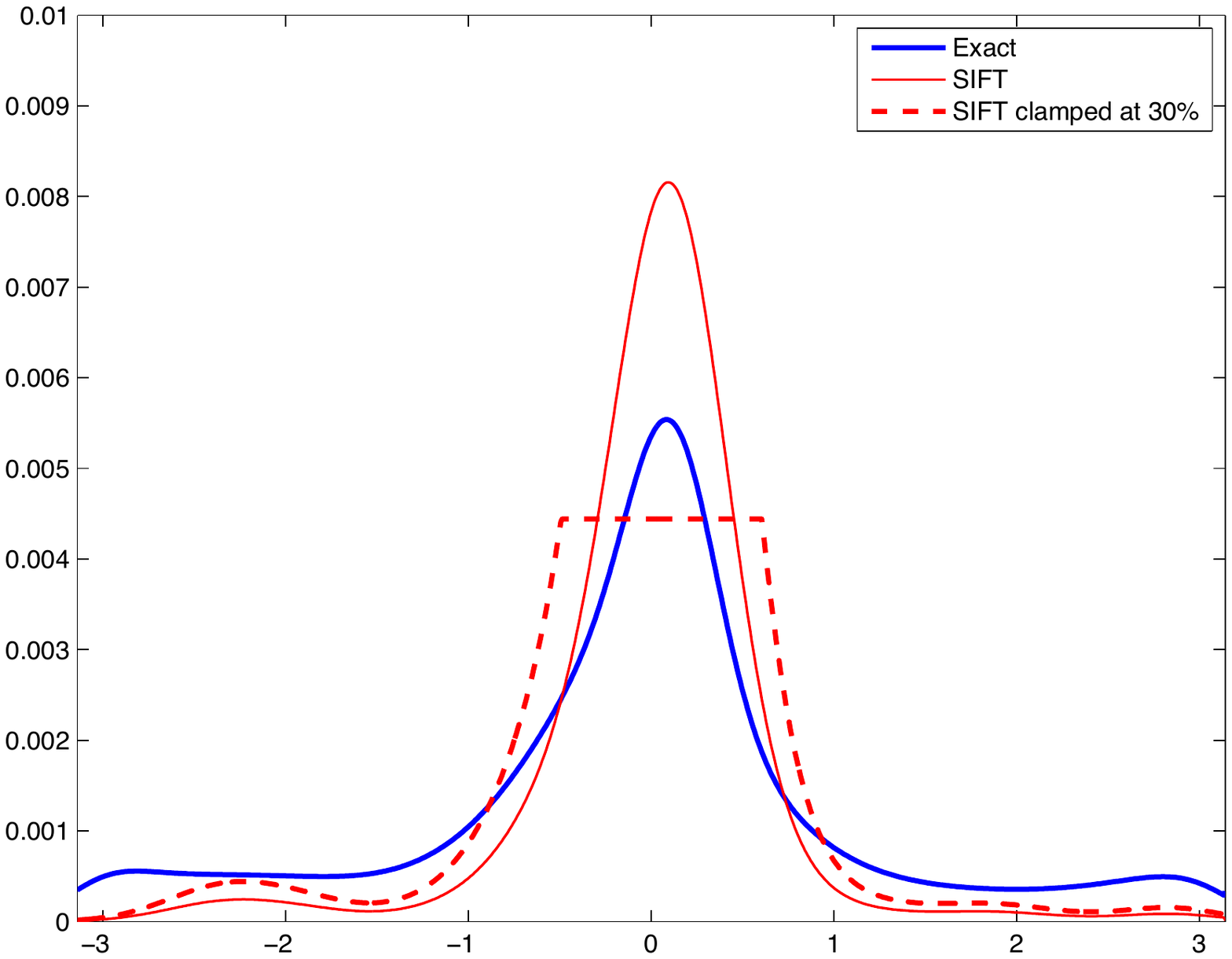}
\includegraphics[height=.1\textwidth,width=.15\textwidth]{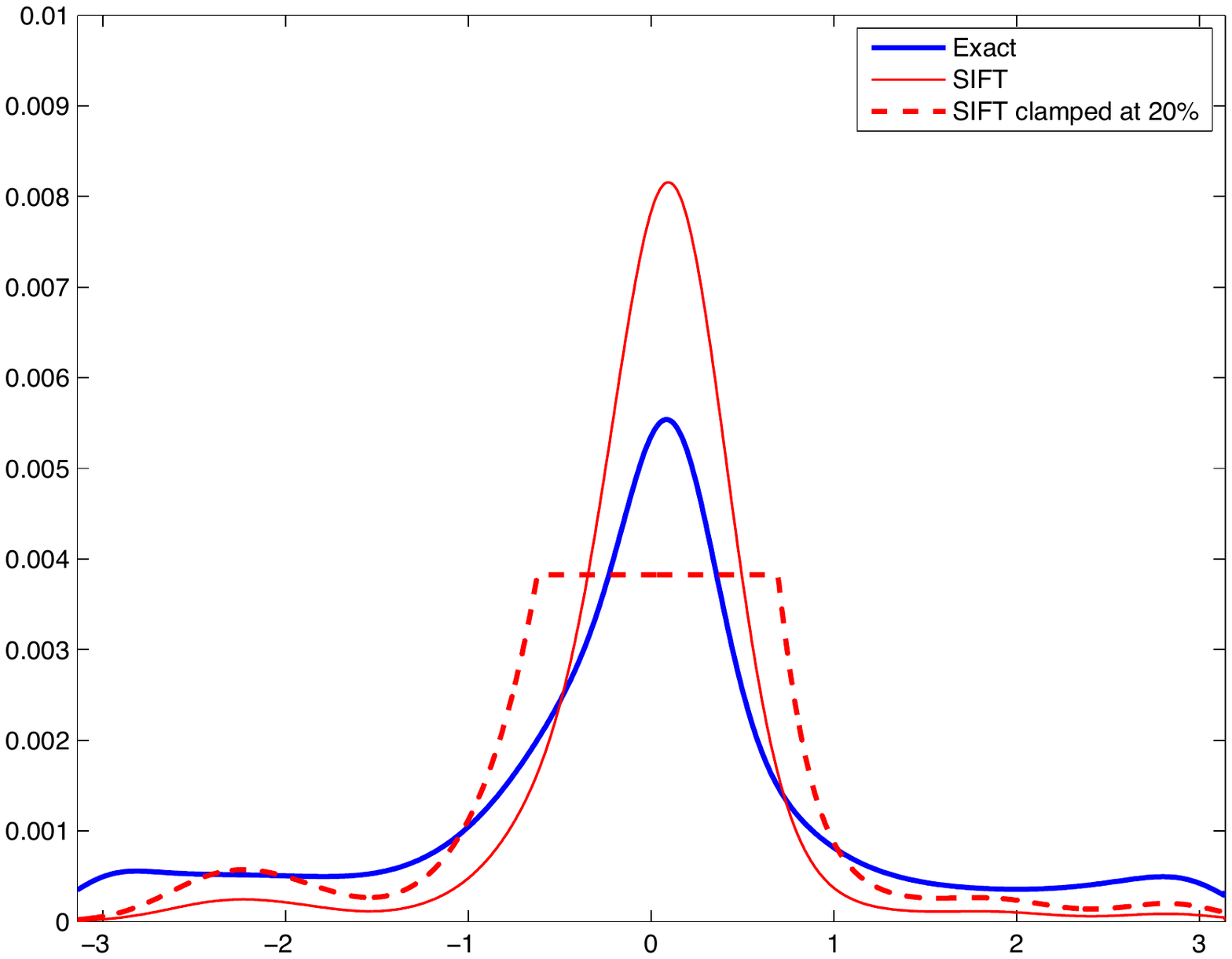}
\includegraphics[height=.1\textwidth,width=.15\textwidth]{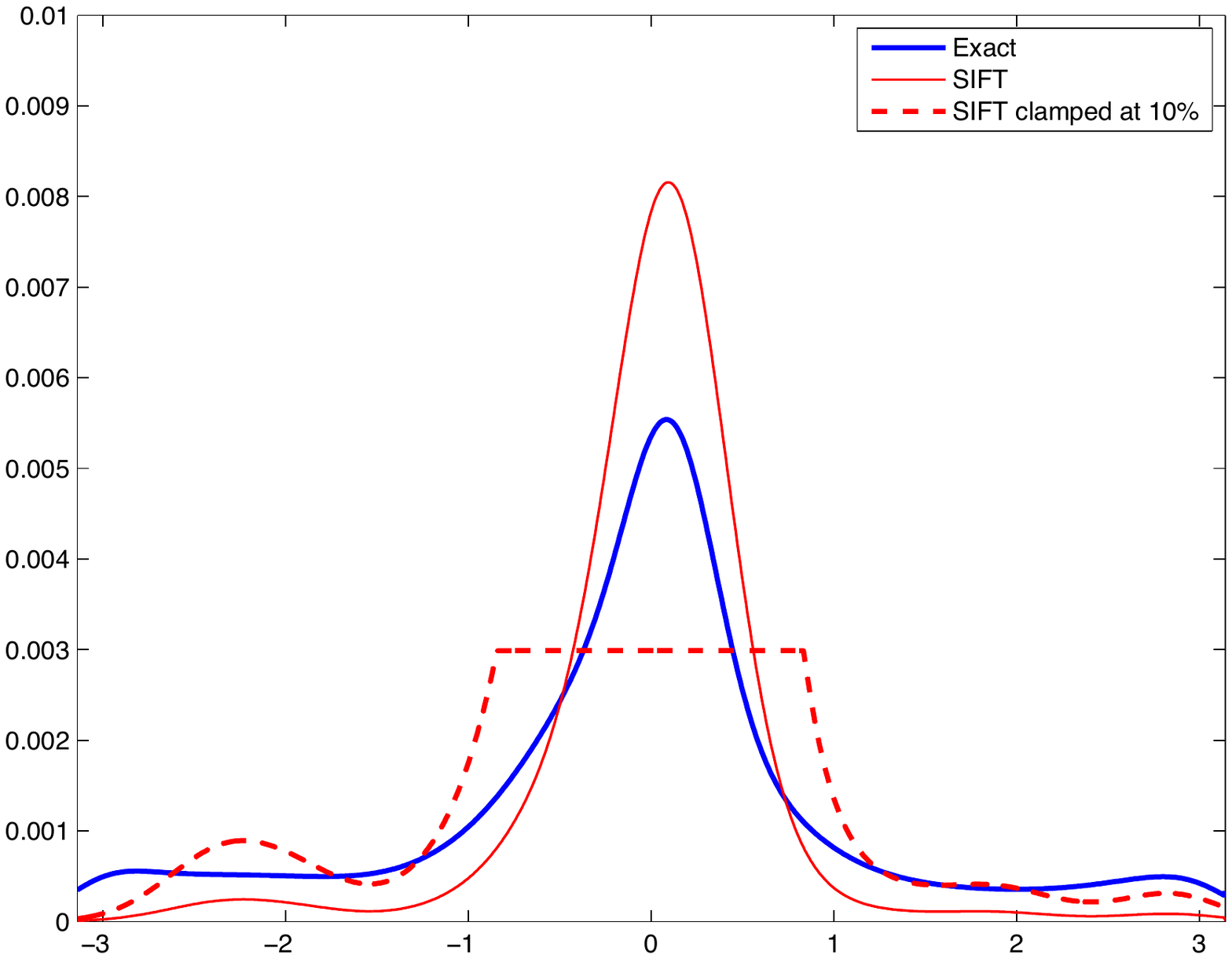}\\
\includegraphics[height=.1\textwidth,width=.15\textwidth]{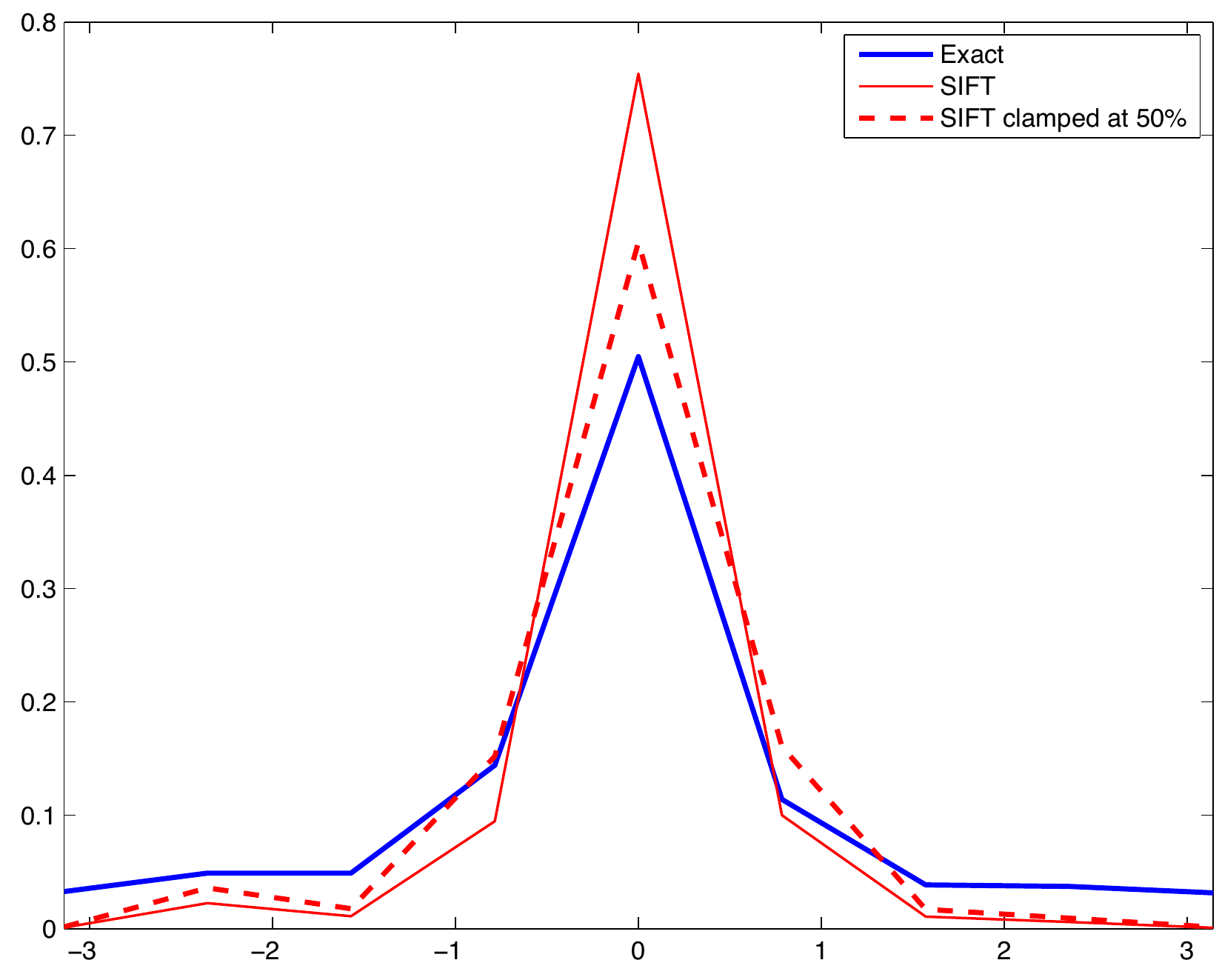}
\includegraphics[height=.1\textwidth,width=.15\textwidth]{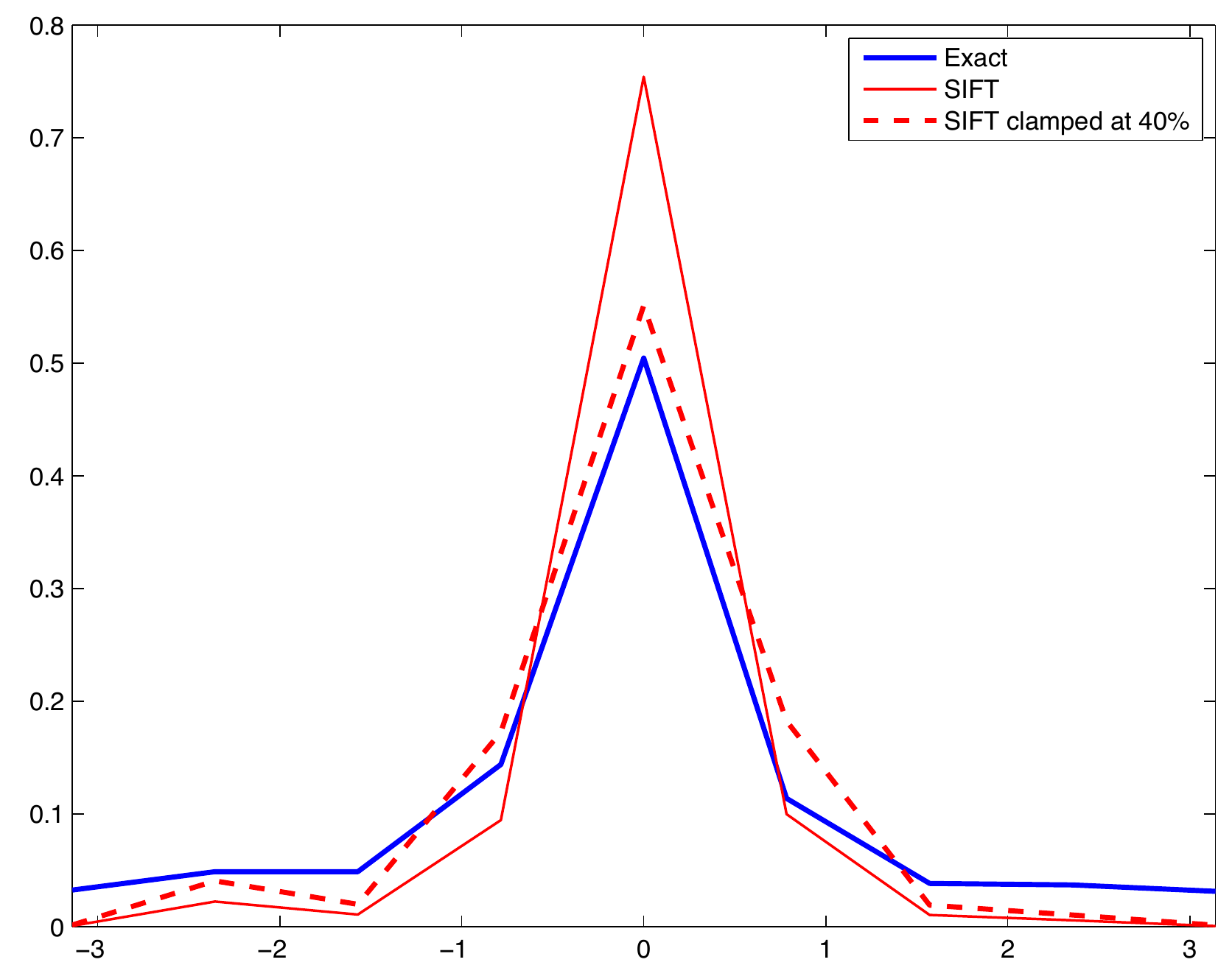}
\includegraphics[height=.1\textwidth,width=.15\textwidth]{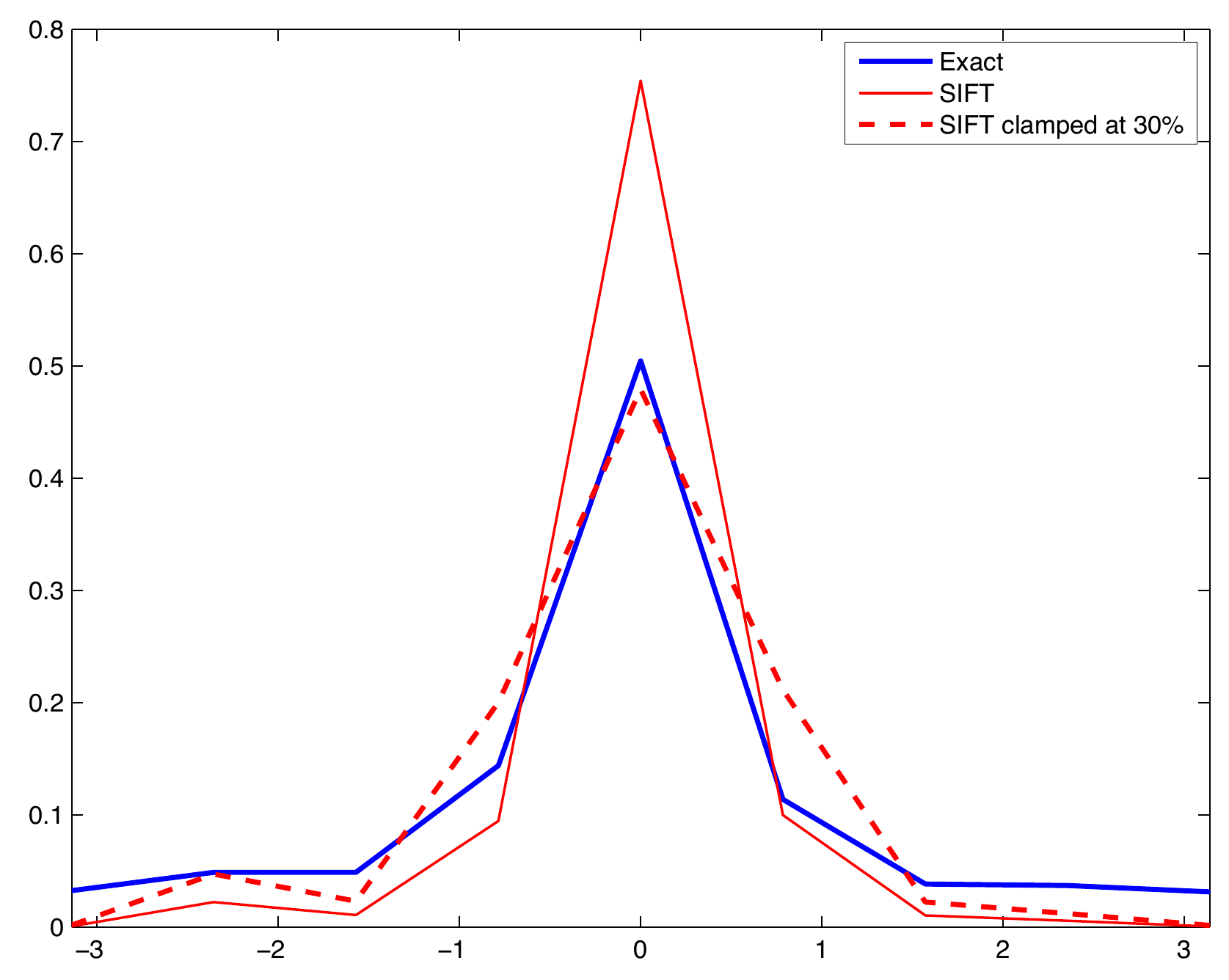}
\includegraphics[height=.1\textwidth,width=.15\textwidth]{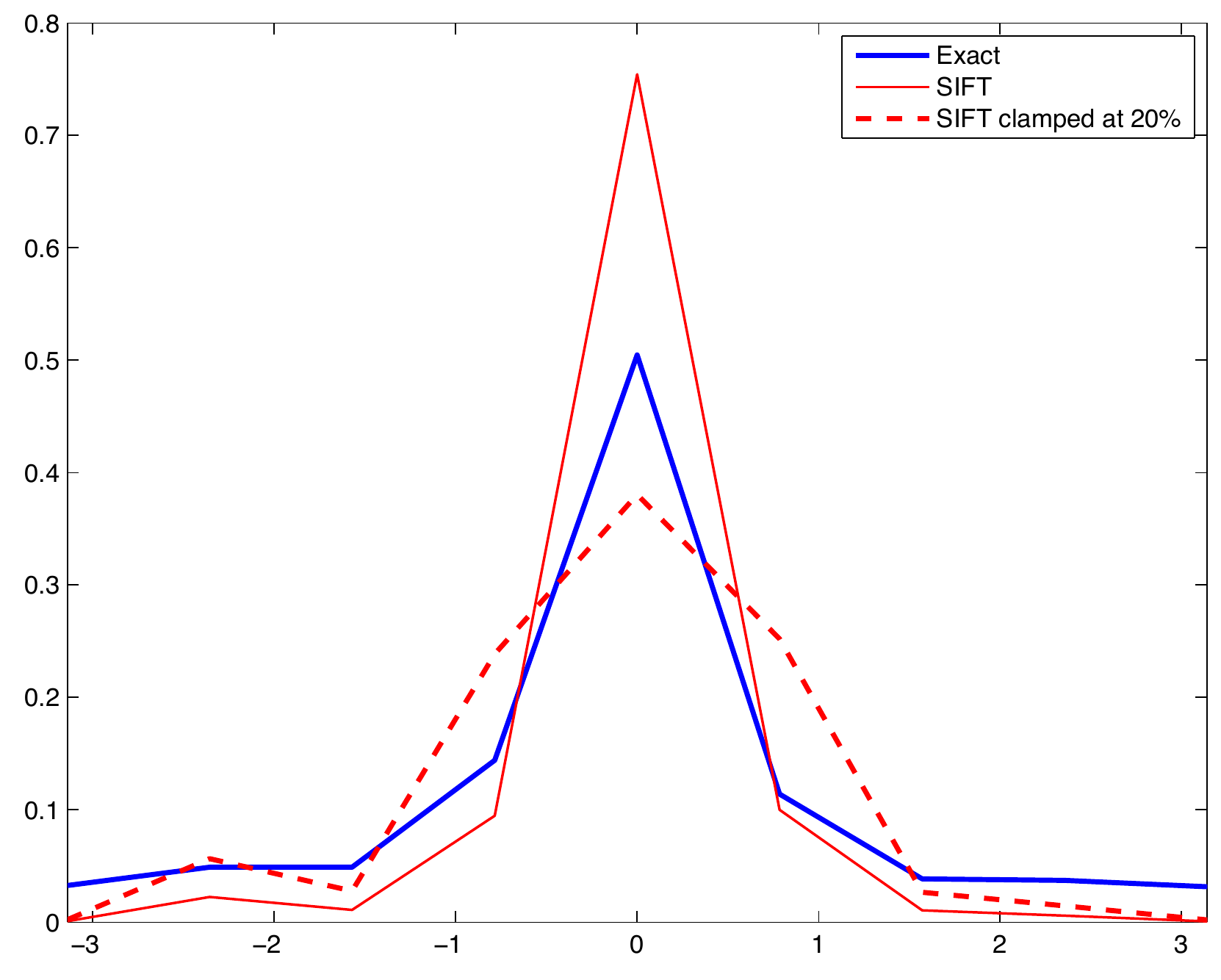}
\includegraphics[height=.1\textwidth,width=.15\textwidth]{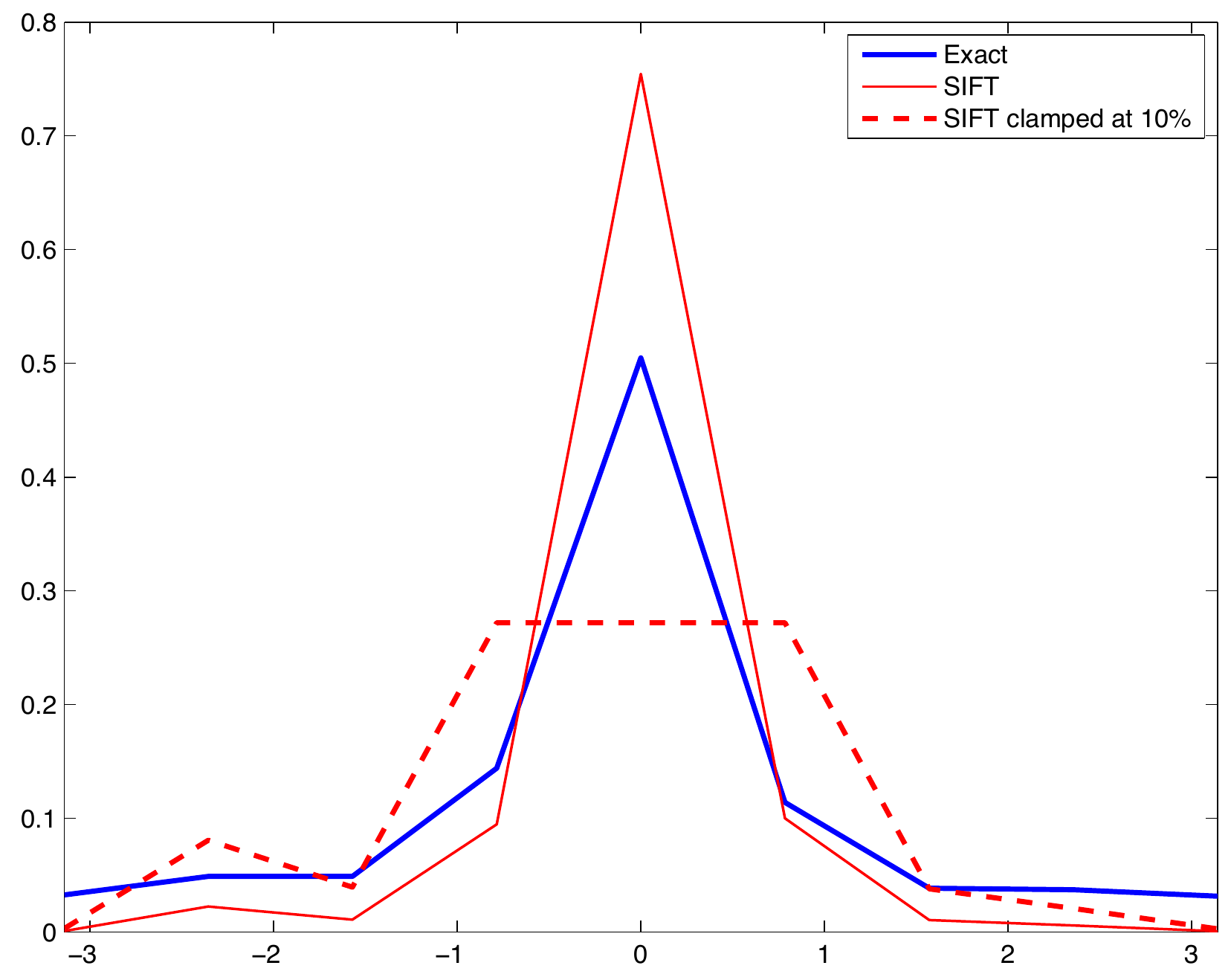}
\end{center}
\vspace{-.5cm}
\caption{\sl {\bf Clamping effects:} For $\alpha \in [-\pi, \pi]$ (abscissa), the top shows the marginalized likelihood $p(\alpha | \nabla \xx)$ \eqref{eq-contr-inv} (blue), the SIFT integrand \eqref{eq-sift-integrand} (solid red), and its clamped version \eqref{eq-clamp} (dashed red) for thresholds ranging from $50\%$ to $10\%$ of its maximum. The bottom shows the same discretized to $8$ orientation bins. The clamping approximation is sensible only for coarse binning, and heavily influenced by the choice of threshold. For an $8$-bin histogram, the best approximation is typically achieved with clamping threshold between $10\%$ and  $30\%$ of the maximum; note that \cite{lowe04distinctive} empirically chose $20\%$.}
\label{fig-clamping}
\end{figure}

Here, we show empirically that \eqref{eq-sift-integrand} becomes closer to  \eqref{eq-contr-inv} after clamping for certain choices of threshold $\tau$ and sufficiently coarse binning. Fig. \ref{fig-clamping} shows that, without clamping, \eqref{eq-sift-integrand} is considerably more peaked than \eqref{eq-contr-inv} and has thinner tails. After clamping, however, the approximation improves, and for coarse binning and threshold between around $20\%$ and $30\%$ the two are very similar. 

\vspacetitle
\subsection{Rotation Invariance}
\label{ex-rotation}

Canonization \cite{soatto09} is particularly well suited to deal with planar rotation, since it is possible to design co-variant detectors with few isolated extrema. An example is the local maximum of the norm of the gradient along the direction $\ttheta = \hat \ttheta_l(\xx)$.\cutForReview{\footnote{Here $g$ acts on $\xx$ via $g\xx(u_i, v_i) = \xx(u_i'', v_i'')$ where $u'' = u \cos \ttheta  -v \sin\ttheta$ and $v'' = u \sin\ttheta  + v \cos \ttheta$, and a canonical element $\hat g_l(\xx) = \hat \ttheta$ can be obtained as $\hat \ttheta = \arg\max_\ttheta \| \nabla \xx(u_i', v_i') \|$. The corresponding rotation invariant $\hat g^{-1}(\xx) \xx$ is $\angle \nabla \xx(u_i', v_i')$ where $u' = u \cos \ttheta  + v \sin\ttheta$ and $v' = - u \sin\ttheta  + v \cos \ttheta \doteq \ttheta'$.}}
Invariance to $G = SO(2)$ can be achieved by retaining the samples
$
\{ p_\M(\ttheta | \hat \ttheta_l) \}_{l = 1}^L.
$
When no consistent (sometimes referred to as ``stable'') reference $\hat \alpha_l$ can be found, it means that there is no co-variant functional with isolated extrema with respect to the rotation group, which means that the data is already invariant to rotation. 

 Note that, again, planar rotations can affect both the training image $\xx$ and the test image $\f$. In some cases, a consistent reference (canonical element) is available. For instance, for geo-referenced scenes $L=1$, and the projection of the gravity vector onto the image plane \cutForReview{\cite{jonesS09}}, $\hat \ttheta$, provides a canonical reference unless the two are orthogonal: 
\be
p_{\M}(\ttheta | G) = p_\M(\ttheta | \hat \ttheta).
\ee

\def\subM{{\xx_{{|}_{\Omega_\xx}}}}
\def\subxx{{\f_{{|}_{\Omega_\f}}}}
\def\N{{{\cal N}_\epsilon}}
\def\xOm{{\f_{|_{\Omega}}}}
\def\xOmb{{\f_{|_{\bar \Omega}}}}
\def\xom{{\f_{|_{\omega}}}}
\def\zOm{{\xx_{|_{\Omega}}}}
\def\zOmb{{\xx_{|_{\bar \Omega}}}}
\def\zom{{\xx_{|_{\omega}}}}

\section{Deep convolutional architectures}
\label{sect-stacking}
\label{sect-nn}

In this section we study the approximation of \eqref{eq-all} implemented by convolutional architectures.\cutForReview{ Note that the representation $\hat p_{\M,G, \hat V}$ was derived from first principles, with no particular attention to biological plausibility or computational feasibility. Here we study approximations of that representation guided by the principles of {\em hierarchy} and {\em compositionality}.} Starting from \eqref{eq-all} for a particular class and a finite number of receptive fields
\cutForReview{\be
\hat p_{\M,G, \hat V}(\f) =  \int_{G^M} \prod_{j \in \hat V} \hat p_{\M ,g_{j}}(\f) 
dP_G(\{g_{j}\} | \M)  
\ee
}
we notice that, since the ``true scene'' $\M$ and the nuisances $g$ are unknown, we cannot factor the likelihood $p_{\M,g}(\f)$ into a product of $p_{\M,g_j}$, which would correspond to a ``bag-of-words'' discarding the dependencies in $dP_G(\{g_j\} |\M)$. 
Convolutional architectures (CNNs) promise to capture such dependencies by hierarchical decomposition into progressively larger receptive fields. Each ``layer'' is a collection of separator (hidden) variables (nodes) that make lower layers (approximately) conditionally independent.\cutForReview{ Then, receptors at each  layer can be interpreted as (single) training images for the children nodes. The relation between the profile likelihood and a convolutional architecture is illustrated next.}

\subsection{Stacking simplifies nuisance marginalization}
\label{sect-stacking-simplifies}

We show this in several steps. We first argue that managing the group of diffeomorphisms can be accomplished by independently managing {\em small} diffeomorphisms in each layer. We use marginalization, but a similar argument can be constructed for max-out or the SA likelihood. Then, we leverage on the local restrictions induced by receptive fields, to deal with occlusion, and argue that such small diffeomorphisms can be reduced locally to a {\em simpler group} (affine, similarity, location-scale or even translations, the most common choice in convolutional architectures). Then global marginalization of diffeomorphisms can be accomplished by local marginalization of the reduced group in each layer. The following lemma establishes that global diffeomorphisms can be approximated by the composition of small diffeomorphisms.
\begin{lemma}
Let $g \in G, \ g: D \rightarrow D$ be an orientation-preserving diffeomorphism of a compact subset $D$ of the real plane, $e \in G$ the identity, and $d(e, g)$ the ``size'' of $g$. Then for any $\epsilon > 0 $ there exists an $N < \infty$ and $g^1, \dots, g^N$ such that $g = g^1 \circ g^2 \dots \circ g^N$ and $d(e, g^i) < \epsilon \ \forall \ i = 1, \dots, N$. 
\end{lemma}
Now for two layers, let $g =  g^1 \circ g^2$, with $g^1, g^2 \sim p(g)$ drawn independently from a prior on $G$. Then $p(g | g^1, g^2) = \delta (g - g^1 \circ g^2)$ (or a non-degenerate distribution if $g^i$ are approximated by elements of the reduced group). Then let $\M^1 = g^2 \M$, or more generally let $\M^1$ be defined such that $\M^1 \perp g^1 \ | \ \M, g^2$. We then have
\bea
p_{{}_G}(\f | \M) &=& \int p(\f | \M, g)dP(g) = \int p(\f | \M^1, g^1,g)dP(\M^1, g^1, g^2 | \M, g) dP(g) = \\
%&=& \int p(\f | \M^1, g^1, g)p(\M^1 | \M, g, g^1, g^2)p(g^1, g^2 | \M, g)dP(g)d\M^1dg^1, dg^2 = \\
&=& \int p(\f | \M^1, g^1)dP(g^1 | \M) p(\M^1 | \M, g^2)dP(g^2 | \M)d\M^1  
\eea
where we have also used the fact that $\f \perp \M \ | \ \M^1$. Once the separator variable $\M^1$ is reduced to a number $K_1$ of filters, we have
\be
p_{{}_G}(\f | \M) \simeq \sum_{k=1}^{K_1} \int p(\f | \M^1_k, g^1)dP(g^1 | \M) \int p(\M^1_k | \M, g^2)dP(g^2| \M) \simeq \sum_{k=1}^{K_1} p_{{}_G}(\f | \M^1_k) p_{{}_G}(\M^1_k | \M) 
\ee
in either case, by extending this construction to $L = N$ layers, we can see that marginalization of each layer can be performed with respect to (conditionally) independent diffeomorphisms that can be chosen to be small per the Lemma above. 
\begin{claim}
Marginalization of the likelihood with respect to an arbitrary diffeomorphism $g\in G$ can be achieved by introducing layers of hidden variables $\M^l$  $l = 1, \dots, L$ and independently marginalizing {\em small} diffeomorphisms $g^l \in G$ at each layer.
\end{claim}
The next step is to restrict the marginalization to each receptive field, at which point it can be approximated by a reduced subgroup, or the (linear) generators.

\subsection{Hierarchical decomposition of the likelihood}
\label{sect-local}

Let 
\be 
p_{{}_G}(\f | \M) \doteq \int_G p(\f | \M, g)dP(g)
\label{eq-marg-lik}
\ee
be the marginal likelihood with respect to some prior on $G$ and introduce a layer of ``separator variables'' $\M^1$ and group actions $g$,  defined such that $\f \perp \M \ | \ (\M^1,g^1)$. This can always be done by choosing $\M^1 = \f$; we will address non-trivial choices later. In either case, forgoing the subscript $G$, 
\be
p(\f | \M) = \int p(\f | \M^1,g^1)dP(\M^1,g^1 | \M).
\ee
If $\M^1$ and $g^1$ take a finite number $K_1$ and $L_1$ of values $\{\M^1_1, \dots, \M^1_{K_1}\}$  (filters) and $\{g^1_1, \dots, g^1_{L_1}\}$, then the above reduces to a sum over $k = 1, \dots, K_1$ and $\ell_1=1, \dots L_1$; the conditional likelihoods   $\{p(\f | \M_1^1,g^1_j), \dots, p(\f | \M^1_{K_1},g^1_j)\}$ are the  {\em feature maps}. If $\f$ has dimensions $N\times M$ and the group actions $g^1_j$ are taken to be pixel wise translations across the image plane, so that $L_1=N\times M$, the feature maps $p(\f | \M^1,g^1)$ can be represented as a tensor with dimensions $N\times M \times K_1$. One can  repeat the procedure for new separator variables that take $K_2$ possible values, and group actions $g^2$ that take $L_2=N_1\times M_1$ values;  the filters $\M^2$ {\em must be supported on the entire feature maps} $p(\f | \M^1,g^1)$ (\ie take values in $N_1\times M_1 \times K_1$) for the sum over $k = 1, \dots, K_1$ to implement the marginalization above 
\be
p(\f | \M) = \sum_{\ell_2=1}^{L_2} \sum_{j=1}^{K_2}\left[\sum_{\ell_1=1}^{L_1}\sum_{k=1}^{K_1} p(\f | \M^1_k,g_{\ell_1}^1) p(\M^1_k,g_{\ell_1}^1 | \M_j^2,g_{\ell_2}^2) \right] p(\M_j^2,g_{\ell_2}^2 | \M).
\ee
The sum is implemented in convolutional networks by the use of translation invariant filters:
\begin{enumerate}
\item At the first layer the support of  $\M^1$ is a small fraction of $N\times M$ and $g^1$ acts on $y$ so that\footnote{There is a non-trivial approximation here, namely that context is neglected when assuming that the likelihood $p(g_{\ell_1}^1 \f | \M^1_k)$  depends only on the restriction of $\f$ to the receptive field $V_j$; see also Section \ref{sect-occlusion} and equation \eqref{eq:nocontext}.} $p(\f | \M^1_k,g_{\ell_1}^1) = p(g_{\ell_1}^1 \f | \M^1_k) = p(\f_{|V_j} | \M^1_k)$.
\item At the second layer the filter $p(\M^1_k,g_{\ell_1}^1 | \M_j^2,g_{\ell_2}^2)$  is nonzero for a finite (and small) number  of group actions $g_{\ell_1}^1$ and also satisfies the shift invariant (convolutional) property
$p(\M^1_k,g_{\ell_1}^1 | \M_j^2,g_{\ell_2}^2) = p(\M^1_k,g_{\ell_2}^2 g_{\ell_1}^1 | \M_j^2)$
\end{enumerate}
 The third dimension of the filters is the number of feature maps in the previous layer.

\def\fj{\f_{|_{V_j}}}

\subsection{Approximation of the first layer}
\label{sect-layer1}

Each node in the first layer computes a local representation \eqref{eq-dsp} using parent node(s) as a ``scene.'' This relates to existing convolutional architectures where nodes compute the response of {\em a rectified linear unit (ReLu) to a filter bank}.\cutForReview{ An optimal representation of a single image $\M = \xx$, computed at location $(u,v) \in \real^2$, marginalized with respect to the location-scale group in a neighborhood ${\cal B}_{\sigma}(u,v)$ is given by \eqref{eq-dsp}.} For simplicity we restrict $G$ to the translation group, thus reducing \eqref{eq-dsp} to SIFT, but the arguments apply to similarities. 

A ReLu response at $(u,v)$ to an oriented filter bank ${\cal G}$ with scale $\sigma$ and orientation $\alpha$ is given by $R_+(\alpha, u, v, \sigma) = \max(0,{\cal G}(u, v; \sigma, \alpha) * \xx(u,v))$. Let ${\cal N}(u, v; \sigma)$ be a Gaussian, centered in $(u,v)$ with isotropic variance $\sigma I$, $\nabla {\cal N}(u,v; \sigma) = [ \frac{\partial {\cal N}}{\partial u}(u,v; \sigma) \ \frac{\partial {\cal N}}{\partial v}(u,v; \sigma)]$, $r(\alpha) = [\cos\alpha \ \sin\alpha ]^T$. Then ${\cal G}(u,v; \sigma, \alpha) \doteq \nabla {\cal N}(u,v; \sigma)r(\alpha)$ is a directional filter with principal orientation $\alpha$ and scale $\sigma$. Omitting rectification for now, the response of an image to a filter bank obtained by varying $\alpha \in [-\pi, \ \pi]$, at each location $(u,v)$ and for all scales $\sigma$ is obtained as 
\bea
R(\alpha, u,v, \sigma) & = & {\cal G}(u,v; \sigma, \alpha) * \xx(u,v) %\\
 %&  &  
%=\nabla {\cal N}(u,v; \sigma)r(\alpha) * \xx(u,v) \\
%&  = &  \nabla {\cal N}(u, v; \sigma) * \xx(u,v) r(\alpha) %\\
% &&
=  {\cal N}(u,v; \sigma) * \nabla \xx(u,v) r(\alpha) \\
 & = & {\cal N}(u,v; \sigma) * \big\langle \frac{ \nabla \xx(u,v)}{\| \nabla \xx(u,v)\|},  r(\alpha) \big\rangle  \| \nabla \xx(u,v)\|\\
%& = & {\cal N}(u,v; \sigma) * \kappa(\angle \nabla \xx(u,v) - \alpha)   \| \nabla \xx(u,v)\| \\ 
&= & \int {\cal N}(u-\tilde u, v-\tilde v; \sigma) \kappa(\angle \nabla \xx(\tilde u,\tilde v), \alpha) \| \nabla \xx(\tilde u, \tilde v) \| d\tilde u d\tilde v
\label{eq-norelu}
\eea
where $\kappa$, the cosine function, has to be rectified for the above to approximate a histogram, 
$
%\be
\kappa_+(\alpha) = \max(0, \cos\alpha) 
%\ee
$ which yields\cutForReview{ a (regularized) approximation of the histogram of orientations of the image gradient in a neighborhood of size $\sigma$ around $(u,v)$, weighted by the gradient norm, {\em i.e,}} SIFT. Unfortunately, in general the latter does not equal\cutForReview{ the ReLu response} $\max(0, {\cal G}*\xx)$ for one cannot simply move the maximum inside the integral. However, under conditions on $\xx$, which are typically satisfied by natural images, this is the case.
\cutForReview{\begin{figure}[htb]
\begin{center}
\includegraphics[width=.25\textwidth]{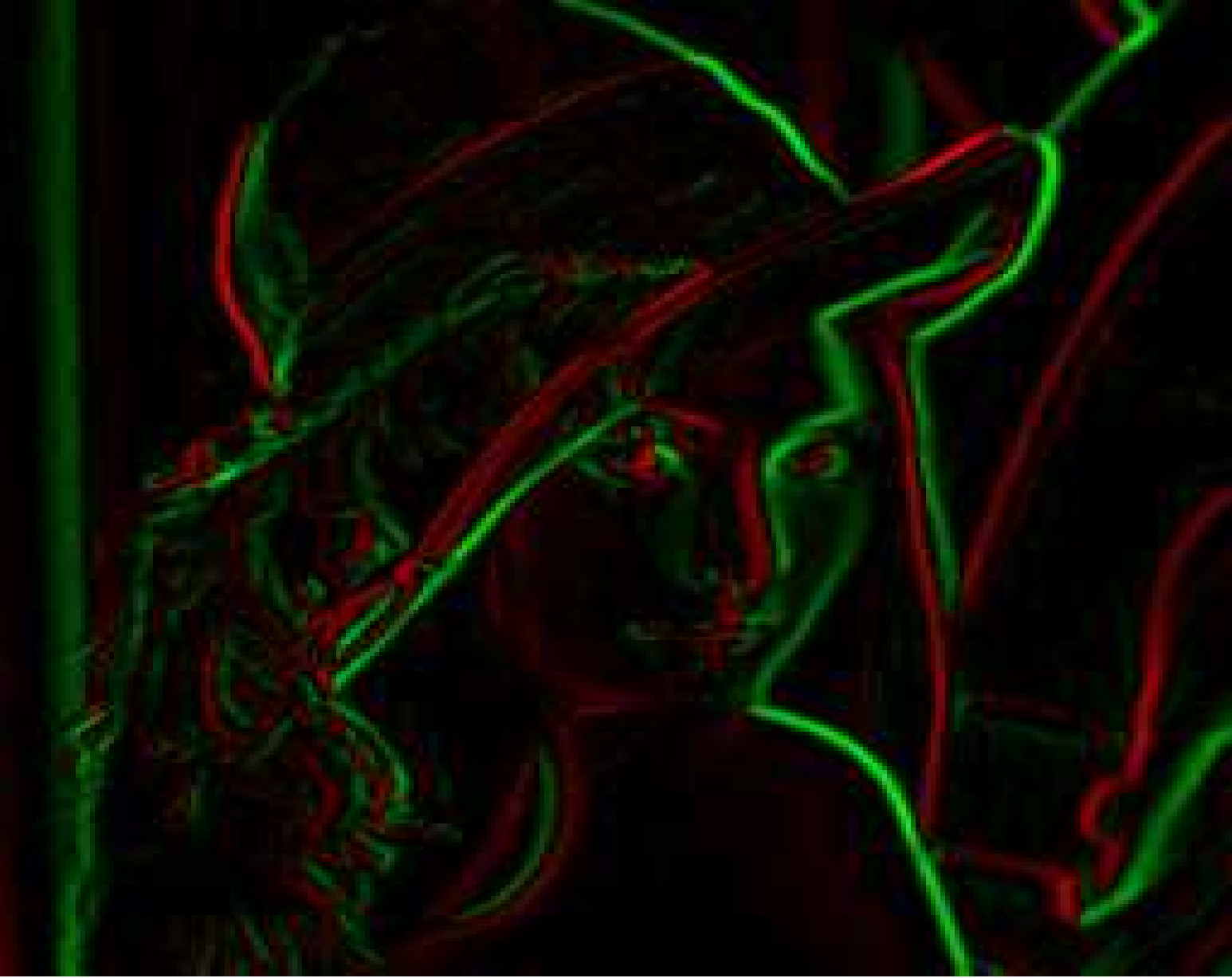}
\end{center}
\caption{\sl \small  $D_+(0)$ (green) and $D_-(0)$, the positive and negative responses to a gradient filter in the horizontal direction. The black region is their complement, which separates them.}
\label{fig-dPlusMinuse}
\end{figure}}
\begin{claim}
\label{claim-1stlayer}
Let $\cal G$ be positive, smooth and have a small effective support $\sigma < \infty$. I.e., $\forall \ \epsilon_1, \epsilon_2 \ \exists \ \sigma  \ | \ {\rm vol}({\cal G}(\tilde u, \tilde v; \sigma, \alpha) \ge \epsilon_1) < \epsilon_2$. Let $\xx$ have a sparse and continuous gradient field, so that for every $\alpha$ the domain of $\xx$ can be partitioned in three (multiply-connected) regions $D_+(\alpha), \ D_-(\alpha)$ and the remainder (the complement of their union), where the projection of the gradient in the direction $\alpha$ is, respectively, positive, negative, and negligible, and $d(\alpha) > 0$ the minimum distance between the regions $D_+$ and $D_-$. Then, provided that $\sigma \le \min_\alpha d(\alpha)$, we have that 
\be
\underbrace{\max(0, {\cal G}(u,v; \sigma, \alpha) * \xx(u,v))}_{\rm ReLu} \simeq  \underbrace{\int {\cal N}(u-\tilde u, v-\tilde v; \sigma) \kappa_+(\angle \nabla \xx(\tilde u,\tilde v), \alpha) \| \nabla \xx(\tilde u, \tilde v) \| d\tilde u d\tilde v}_{\rm sift}
\ee
\end{claim}

\subsection{Stacking information} 

A local hierarchical architecture allows approximating the SA likelihood $p_{\M, G}(\cdot)$ by reducing nuisance management to local marginalization and max-out of simple group transformations. The SA likelihood $p_{\M, G}(\f)$ is an optimal representation for any query on $\M$ given data $\f$. For instance, for classification, the representation $p_{\M, G}(\f)$ is itself the classifier (discriminant). Thus, if we could compute an optimal classifier, the representation would be the identity; vice-versa, if we could compute the optimal representation, the classifier would be a threshold. In practice, one restricts the family of classifiers -- for instance to soft-max, or SVM, or linear -- leaving the job of feeding the most informative statistic to the classifier. In a hierarchical architecture, this is the feature maps in the last layer. This is equivalent to neglecting the global dependency $p_{\M, G}(\f | \M^L)$ on $\M$ at the last layer. The information loss inherent in this choice is the loss of assuming that $\M^L$ are independent (whereas they are only independent conditioned on $\M$).

An optimal representation with restricted complexity $L < \infty$, therefore, maximizes the independence of the components of $\M^L$, or equivalently the independence of the components of $\f$ given $\M^L$. %This establishes a link between the information content of a representation \eqref{eq-actinf} in Sec. \ref{sect-likelihood} and \cite{ver2015maximally}. 
Using those results, one can show that the information content of a representation (\eqref{eq-actinf} in App. \ref{sect-likelihood}) grows with the number of layers $L$. 

\vspacetitle
\section{Discussion}
\label{sect-discussion}
\vspacetitle

For the likelihood interpretation of a CNN put forward here to make sense, training should be performed {\em generatively}, so fixing the class $\M_k$ one could sample (hallucinate) future images $\f$ from the class. However neither the architecture not the training of current CNN incorporate mechanisms to enforce the statistics of natural images.

In this paper we emphasize the role of the {\em task} in the representation: If nothing is known about the task, nothing interesting can be said about the representation, and the only optimal one is the trivial one that stores all the data. This is because the task could end up being a query about the value of a particular pixel in a particular image. Nevertheless, there may be many different tasks that share the same representation by virtue of the fact that they share the same nuisances. In fact, the task affects what are nuisance factors, and the nuisance factors affect the design and learning of the representation. For some complex tasks, writing the likelihood function may be prohibitively complex, but some classes of nuisances such as changes of illumination or occlusions, are common to many tasks. 

Note that, by definition, a nuisance is not informative. Certain transformations that are nuisances for some tasks may be informative for others (and therefore would not be called nuisances). For instance, viewpoint is a nuisance in object detection tasks, as we want to detect objects regardless of where they are. It is, of course, not a nuisance for navigation, as we must control our pose relative to the surrounding environment. In some cases, nuisance and intrinsic variability can be entangled, as for the case of intra-class deformations and viewpoint-induced deformations in object categorization. Nevertheless, the deformation would be informative {\em if it was known or measured}, but since it is not, it must be marginalized. 

Our framework does not require nuisances to have the structure of a group. For instance, occlusions do not. Invariance and sensitivity are still defined, as a statistic is invariant if it is constant with respect to variations of the nuisance. What is not defined is the notion of {\em maximal invariance}, that requires the orbit structure. However, in our theory maximal invariance is not the focus. Instead, {\em sufficient invariance} is. 

The literature on the topic of representation is vast and growing. We focus on visual representations, where several have been active. \cite{anselmi2015invariance} have developed a theory of representation aiming at approximating maximal invariants, which restricts nuisances to (locally) compact groups and therefore do not explicitly handle occlusions. Both frameworks achieve invariance at the expense of discriminative power, whereas in our framework both can be attained at the cost of complexity. \cite{patel2015probabilistic}, that appeared after earlier drafts of this manuscript were made public, instead of of starting from principles and showing that they lead to a particular kind of computational architecture, instead assume a particular architecture and interpret it probabilistically, similarly to \cite{ver2015maximally} that uses total correlation as a proxy of information, which is related to our App. \ref{sect-likelihood}. However, there the representation is defined absent a task, so the analysis does not account for the role of nuisance factors.  

In particular, \cite{anselmi2015invariance} define a ${\cal G}$-invariant representation $\mu$ of a (single) image $I$ as being {\em selective} if $\mu(I) = \mu(I') \Rightarrow I \sim I'$ for all $I, I'$, \ie if it is a {\em maximal invariant}. But while equivalence to the data up to the action of the nuisance group is  critical for {\em reconstruction}, it is not necessary for other tasks. Furthermore, for non-group nuisances, such as occlusions, a maximal invariant cannot be constructed. Instead, given a task, we replace maximality with {\em sufficiency} for the task, and define at the outset an optimal representation to be a {\em minimal sufficient invariant statistic,} or ``sufficient invariant,'' approximated by the {\em SAL Likelihood.} The construction in \cite{anselmi2015invariance} guarantees maximality for compact groups. Similarly, \cite{sundaramoorthiPVS09} have shown that maximal invariants can be constructed even for diffeomorphisms, which are infinite-dimensional and non-compact. In practice, however, {\em occlusions and scaling/quantization} break the group structure, and therefore a different approach is needed that relies on sufficiency, not maximality, as we proposed here. To relate our approach to \cite{anselmi2015invariance}, we note that the orbit probability of Def. 4.2 is given by 
\be
\rho_I(A) = P(\{g \ | \ g I \in A\})
\ee
and is used to induce a probability on $I$, via $P(I)[A] = P(\{g \ | \ gI \in A\})$. On the other hand, we define the minimal sufficient invariant statistic as the marginalized likelihood 
\be
p_{\M, G}(\f) \doteq \int p_{\M,g}(\f)dP(g)
\ee
where $\f$ is a (future) image, and $\M$ is the {\em scene}. If we consider the scene to be comprised of a set of images $A = \theta$, and the future image $\f = I$, then we see that the OP is a marginalized likelihood where $dP(g) = d\mu(g)$ is the Haar measure, and $p_{\M, g}(\f) = \delta(g\f \cap \M)$. Thus, substitutions $G \leftarrow {\cal G}$, $\theta \leftarrow A$, $\f \leftarrow I$ yield
\be
P(I)[A] = p_{\M,G}(\f)
\ee
for the particular choice of Haar measure and impulsive density $p_{\M, g}$. 
The TP representation can also be understood as a marginalized likelihood, as $\Psi(I)[A]$ is the ${\cal G}$-marginalized likelihood of $I$ given $A$ when using the uniform prior and an impulsive conditional $p_{A, g}(I)$:
\be
\Psi(I)[A] = \int_{\cal G} p(gI | A)d\mu(g).
\ee
Finally, our treatment of representations is not biologically motivated, in the sense that we simply define optimal representations from first principles, without regards for whether they are implementable with biological hardware. However, we have established connections with both local descriptors and deep neural networks, that were derived using biological inspiration.

\subsubsection*{Acknowledgments} Work supported by FA9550-15-1-0229, ARO W911NF-15-1-0564, ONR N00014-15-1-2261, and MIUR RBFR12M3AC ``Learning meets time.''  We are appreciative of discussions with Tomaso Poggio, Lorenzo Rosasco, Stephane Mallat, and Andrea Censi.

%% {\small
%% \itemsep=-1pt
%% \vspacetitle
%% %\bibliographystyle{plain}
%% \bibliographystyle{iclr2016_conference}
%% \bibliography{/Users/soatto/lib/tex/self,/Users/soatto/lib/tex/total2}
%% \vspacetitle
%% }

\newpage
\appendix

\vspacetitle
\section{Quantifying the information content of a representation} 
\label{sect-likelihood}
\vspacetitle

\cutForReview{\begin{figure}[htb]
\begin{center}
\includegraphics[width=.25\textwidth]{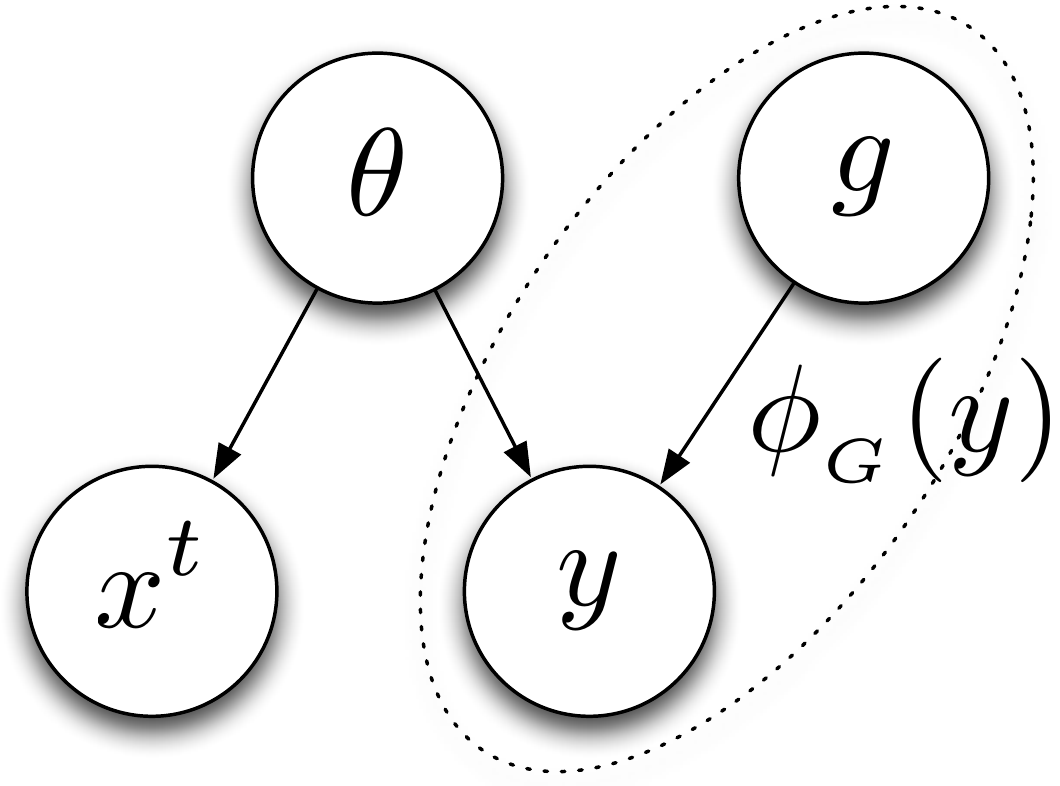}
\end{center}
\caption{\sl \small  Independence structure in the LA model.}
\label{fig-ind}
\end{figure}}

\cutForReview{We are now ready to formalize the notion of representation stated in Sect. \ref{sect-desiderata}: If we had an expression for the SAL likelihood $\hat p_{\M, G, V}(\cdot)$, then as soon as a test datum $\f$ was available, we could answer any question using $\hat p_{\M, G, V}(\f)$.} In general we do not know the likelihood, but\cutForReview{ under the assumptions of the LA model, the scene {\em separates} training and test data\footnote{$\xx^t \perp \f \ | \ \M$, meaning that $p_\M(\xx^t, \f) = p_\M(\xx^t)p_\M(\f)$.} (Fig. \ref{fig-ind}), so what} we have\cutForReview{ is} a collection of {\em samples} $\xx_t \sim p_{\M, g_t}(\xx)$, each generated with some nuisance $g_t$, which we can use to infer, or ``learn'' an approximation of the SAL likelihood \cite{fraser2007minimal}. \cutForReview{Since the set of questions about the scene can be identified with a partition of $\M$, assuming a supervised sample $X^t$ is available corresponding to this partition, then we can approximate the SAL likelihood leading to a learned representation: }
\bea
\phi_\M(\cdot) \doteq p_\M(\cdot)  &\simeq&  \hat p_{{}_{X^t}}(\cdot),  \ X^t \sim p_\M(\cdot) ~~~~~~~~~~~~ {\rm (empirical \ likelihood)} \\
\phi_{\M, G}(\cdot)  \doteq p_{\M, G}(\cdot) &\simeq& \hat p_{{}_{X^t,G}}(\cdot)
% \int\hat p_{{}_{X^t}}(g~ \cdot)dP(g) 
%{\color{red} ?= \hat p_{{}_{X^t}}(\phi_G(\cdot)) ?= \hat p_{{}_{X^t,G}}(\cdot) ?= \phi_G(\cdot)}
~~~~~~~~~~~~~~~~~~~~~~~~~~~~ {\rm (profile \ likelihood)} \\
\phi_{\M, G}(\f,\xx^t) \doteq \phi_{\M, G}(y) \phi_\M(\xx^t) &\simeq&  %\int\hat p_{{}_{X^t}}(\f | g)dP(g)  \hat p_{{}_{X^t}}(\xx^t) \\ 
%& \propto&  \int\hat p_{{}_{X^t}}(\f | g)dP(g)  \doteq  
\hat p_{{}_{X^t, G}}(\f)\hat p_{{}_{X^t}}(\xx^t) \propto \hat p_{{}_{x^t, G}}(\f) % {\color{red} ?= \phi_G(\f) } 
~ {\rm (learned \ representation)} 
\label{repr-def}
\eea

Depending on modeling choices made, including the number of samples $N$, the sampling mechanism, and the priors for local marginalization, the resulting representation will be ``lossy'' compared to the data. Next we quantify such a loss.

\subsection{Informative content of a representation}
\label{sect-info-content}

The scene $\M$ is in general infinite-dimensional. Thus, the informative content of the data cannot be directly quantified by mutual information ${\mathbb I}(\M; \xx^t)$. In fact, ${\mathbb H}(\M) = \infty$ and therefore, no matter how many (finite) data $\xx^t$ we have,  ${\mathbb I}(\M; \xx^t) = {\mathbb H}(\M) - {\mathbb H}(\M | \xx^t) = \infty - \infty$ is not defined. Similarly, ${\mathbb I}(\M; \phi(\xx^t))$ is undefined and therefore mutual information cannot be used directly to measure the informative content of a representation, or to infer the most informative statistic. 

The notion of {\em Actionable Information} \cite{soatto09} as ${\cal H}(\xx^t) \doteq {\mathbb H}(\phi_{\M,G}(\xx^t))$ where $\phi_{\M,G}$ is a maximal $G$-invariant, can be used to bypass the computation of ${\mathbb H}(\M)$: Writing formally  
\be
{\mathbb I}(\M; \phi_{\M,G}(\f)) = {\mathbb H}(\phi_{\M,G}(\f)) - {\mathbb H}(\phi_{\M,G}(\f) | \M)
= {\cal H}(\xx^t) - {\mathbb H}(\phi_{\M,G}(\f) | \M)
\label{eq-actinf}
\ee
we see that, if we were given the scene $\M$, we could generate the data $\f$, under the action of some nuisance $g$, up to the residual modeling uncertainty, which is assumed white and  zero-mean (lest the mean and correlations of the residual can be included in the model). Similarly, we can generate a maximal invariant $\phi_{\M,G}(\f)$ up to a residual with constant entropy; therefore, the statistic which maximizes ${\mathbb I}(\M; \phi_{\M,G}(\f))$ is the one that maximizes the first term in \eqref{eq-actinf}, ${\mathbb H}(\phi_{\M,G}(\f))$. This formal argument allows defining the most informative statistic as the one that maximizes Actionable Information, $\hat \phi_{\M,G} = \arg\max_{\phi_{\M,G}} {\mathbb H}(\phi_{\M,G}(\xx^t)) \doteq {\cal H}(\xx^t)$, bypassing the computation of the entropy of the ``scene'' $\M$. Note that the task still influences the information content of a representation, by defining what are the nuisances $G$, which in turn affect the computation of actionable information.

An alternative approach is to measure the informative content of a representation bypassing consideration of the scene is described next. 
\begin{defn}[Informative Content of a Representation]
The information a statistic $\phi$ of $\xx^t$ conveys on $\M$ is the information it conveys on a task $T$ (\eg a question on the scene $\M$), regardless of nuisances $g\in G$: 
\be
{\mathbb I}(gT; \phi(\xx^t)) = {\mathbb H}(g T) - {\mathbb H}(gT | \phi(\xx^t)) ~~~ \forall \ g\in G
\ee
\label{def-info}
\end{defn}
If the task is {\em reconstruction} (or prediction) $T = \f$, where past data $\xx^t$ and future data $\f$ are generated by the same scene $\M$, then the definition above relates to the past-future mutual information \cite{creutzigGT09} except for the role of the nuisance $g$. The following claim shows that an ideal representation, as previously defined in terms of minimal sufficient invariant statistic, maximizes information.
\begin{claim} Let past data $\xx^t$ and future data $\f$, used to accomplish a task $T$, be generated by the same scene $\M$. Then the representation $\phi_{\M, G}$ maximizes the informative content of a representation. \label{thm-ideal}
\end{claim}
\cutForReview{\begin{proof}
Since $p_{\M,G}(\f,\xx^t)$ is sufficient for $\M$, and it factorizes into $\phi_{\M,G}(\f)\phi_\M(\xx^t)$, then $\phi_\M(\xx^t)$ is sufficient of $\xx^t$ for $\M$. By the factorization theorem (Theorem 3.1 of \cite{pawitan}), there exist functions $f_\M$ and $\psi$ such that $\phi_\M(\xx^t) \propto f_\M(\psi(\xx^t))$, \ie the likelihood depends on the data only through the function $\psi(\cdot)$. This latter function is what is more commonly known as the sufficient statistic, which in particular has the property that $p(\M| \xx^t) = p(\M | \psi(\xx^t))$. However, if $\phi_\M$ is sufficient for $\M$, it is also sufficient for future data generated from $\M$. Formally, 
\be
p_G(\f | \xx^t) = \int p_G(\f | \xx^t, \M)p(\M | \xx^t)dP(\M) = \int p_G(\f | \M)p(\M | \psi(\xx^t)) dP(\M) = p_G(\f | \psi(\xx^t))
\label{eq-proof-inf}
\ee
which shows that $\psi$ minimizes the uncertainty of $\f$ for any $g\in G$ since the right-hand-side is $G$-invariant. 
The right-hand side above is the {\em predictive likelihood}\cutForReview{\footnote{Definition 2 of \cite{hinkley1979predictive} with $Y = \xx^t, Z = \f, S = \phi(\xx^t), T = \phi_G(\f)$ and $R = \phi(\xx^t, \f)$, up to a factor $\frac{p(\phi(\xx^t, \f))}{p(\phi(\xx^t))}$.}}  \cite{hinkley1979predictive}, which must therefore be proportional to $\tilde f_\f(\psi(\xx^t))$ for some $\tilde f$ {\em and the same} $\psi$, also by the factorization theorem. 
\end{proof}}
\vspace{-.1cm}
\cutForReview{\begin{cor} An optimal representation is minimal sufficient for $\M$ and sufficient (but not necessarily minimal) for $\f$; the predictive likelihood is minimal sufficient for $\f$ but not necessarily sufficient for $\M$.
\end{cor}}
The next claim relates a representation to Actionable Information.
\begin{claim}
If $\phi$ maximizes Actionable Information, it also maximizes the informative content of a representation. 
\end{claim}
\cutForReview{\begin{proof} Follows from \eqref{eq-proof-inf} after noticing that, if $\phi$ maximizes Actionable Information, then by definition $\phi$ is a maximal $G$-invariant.
\end{proof}}
Note that maximizing Actionable Information is a stronger condition than maximizing the information content of a representation. Since Actionable Information concerns maximal invariance, sufficiency is automatically implied, and the only role the task plays is the definition of the nuisance group $G$. This is limiting, since we want to handle nuisances that do not have the structure of a group, and therefore Definition \ref{def-info} affords more flexibility than \cite{soatto09}.
\cutForReview{
\begin{rem}[Active Learning]
A representation, informative as it may be, can be no more informative than the data itself (Data Processing Inequality), uninformative as it may be. This is obvious but irrelevant in the context discussed so far, for we have sought statistics that are  {\em as informative as the (training) data}, however good or bad that is. For the representation to (asymptotically) approach the informative content of the scene, it is necessary to {\em design the experiment} $E$ so that the data collected $\xx^t$, with $t \rightarrow \infty$, yields statistics that are asymptotically {\em complete}: Formally, $\lim_{t\rightarrow \infty} {\mathbb H}(\phi_G(\f) | \phi(\xx^t)) = 0$ where the limit is taken with respect to some action $u_t$ that has as a result the collection of data \cite{fedorov1972theory}, would be equivalent to knowing the scene. Such active learning or active sensing is beyond our scope here. 
\end{rem}
}

The next two claims characterize the maximal properties of the profile likelihood. We first recall that the marginalized likelihood is invariant {\em only if} marginalization is done with respect to the base (Haar) measure, and in general it is not a maximal invariant, as one can show easily with a counter-example (\eg a uniform density). On the other hand, the profile likelihood is by construction invariant regardless of the distribution on $G$, but is also -- in general -- not maximal. However, it is maximal under general conditions on the likelihood. To see this, consider $p_\M(\cdot)$ to be given; for a free variable $\xx$, it can be written as a map $q$: $p_\M(\xx) \doteq q(\M, \xx)$.
For a fixed $\xx$, $q$ is a function of $\M$. If $q$ is constant along $\xx$ (the level curves are straight lines), then in general $q(\M, \f) = q(\M, \xx)$ for all $\M$ does {\em not} imply $\f = \xx$. Indeed, $\f$ can be arbitrarily different from $\xx$. Even if $q$ is non-degenerate (non-constant along certain directions), but presents {\em symmetries}, it is possible for different $\f \neq \xx$ to yield the same $q(\M, \f) = q(\M, \xx)$ for all $\M$. However, under {\em generic conditions} $q(\M, \f) = q(\M, \xx)$ for all $\M$ implies $\f = \xx$. 
Now consider $p_{\M,G}(\cdot)$ to be given; for a free variable $\xx$ the map can be written using a function $q$ such that  $p_{\M,G}(\xx) = \min_g q(\M, g\xx)$. Note that $p_{\M, G}$ is, by construction, invariant to $G$. Also note that, following the same argument as above, the invariant is not maximal, for it is possible for $p_{\M, G}(\xx) = p_{\M, G}(\f)$ for all $\M$ and yet $\xx$ and $\f$ are not equivalent (equal up to a constant $g$: $\f = g\xx$). However, if the function $q$ is {\em generic}, then the invariant is maximal. In fact, let $\hat g(\M, \xx) = \arg\min q(\M, g\xx)$, so that $p_{\M, G}(\xx) = q(\M, \hat g(\M, \xx)\xx)$. If we now have $ q(\M, \hat g(\M, \f)\f) =  q(\M, \hat g(\M, \xx)\xx)$ for all $\M$, then we can conclude, based on the argument above, that $\hat g(\M, \xx)\xx = \hat g(\M, \f)\f$. Since $\xx$ and $\f$ are fixed, and the equality is for all $\M$, we can conclude that $\hat g(\M,\f)^{-1}\hat g(\M,\xx)$ is independent of $\M$. That allows us to conclude the following.
\begin{claim}[Maximality of the profile likelihood]
If the density $p_\M(\xx)$ is generic  with respect to $\xx$, then $p_{\M, G}(\cdot)$ is a maximal $G$-invariant.
\end{claim}
Since we do not have control on the function $p_{\M, G}(\cdot)$, which is instead in general constructed from data, it is legitimate to ask what happens when the generic condition above is not satisfied. Fortunately, distributions that yield non-maximal invariants can be ruled out as uninformative at the outset:
\begin{claim}[Non-maximality and non-informativeness]
If $q$ is such that, for any $\xx \neq \f$ we have $q(\M, \hat g(\M, \xx)\xx) = q(\M, \hat g(\M, \f)\f)$ for all $\M$, then $q_{\M, G}(\cdot)$ is uninformative. 
\end{claim}
This follows from the definition of information, for any statistic $T$. 

As we have pointed out before, what matters is not that the invariant be {\em maximal}, but that it be {\em sufficient}. As anticipated in Rem. \ref{rem-sufficient-invariant}, we can achieve invariance with {\em no sacrifice} of discriminative power, albeit at the cost of complexity.

\section{Proofs}

{\bf Theorem \ref{thm-lik}}
\begin{proof} 
Pick any $\M_0$, define $\T(\xx) \doteq \frac{L(\cdot; \xx)}{L(\M_0; \xx)}$ and $f(T(\xx), \M) \doteq \frac{L(\M; \xx)}{L(\M_0; \xx)}$ and apply the factorization lemma with $h(\xx) = L(\M_0; \xx)$. 
\end{proof}

{\bf Theorem \ref{claim-contrast}}\\

\begin{proof}
We denote with $\overline{\nabla \f} \doteq \frac{\nabla \f}{\| \nabla \f \|}$ the normalized gradient of $\f$, and similarly for $\xx$; $\Phi$ maps it to polar coordinates $(\ttheta, \rho) = \Phi(\nabla \f)$ and $(\beta, \gamma)=\Phi(\nabla \xx)$, where 
$$
 \ttheta \doteq \angle \nabla \f \quad \rho \doteq \|\nabla \f \| \quad 
\beta \doteq\angle \nabla \xx \quad
\gamma \doteq \|\nabla \xx\|.
$$
The conditional density of  $\nabla \f$  given $\nabla \xx$ takes the polar form
\be
\label{eq-cond-polar}
\begin{array}{rcl}
p(\rho,\ttheta|\nabla\xx) &=& p(\nabla\f|\nabla\xx)_{\nabla\f=\Phi^{-1}(\rho,\ttheta)} \rho\\
&& p(\nabla\f|\nabla\xx)  = \frac{1}{%\left(\sqrt{
2\pi\epsilon^2
%}\right)^2
} e^{-\frac{1}{2\epsilon^2}\| \nabla\f-\nabla\xx\|^2}.
\end{array}
\ee
Defining $(\nabla\xx)_i$ to be the $i$-th component of $\nabla\xx$, \eqref{eq-cond-polar} can be expanded as
\be\label{eq-polar-polar}
p(\rho,\ttheta| \nabla\xx) = \rho\frac{1}{%\left(\sqrt{
2\pi\epsilon^2
%}\right)^2
} e^{-\frac{1}{2\epsilon^2}\left[ (\rho \cos(\ttheta) - (\nabla\xx)_1)^2 +  (\rho \sin(\ttheta) - (\nabla\xx)_2)^2\right]}
\ee
and the exponent is 
\be\label{eq-exponent}
\begin{array}{rcl}
(\rho \cos(\ttheta) - (\nabla\xx)_1)^2 +  (\rho \sin(\ttheta) - (\nabla\xx)_2)^2 & = & \rho^2 - 2\rho ( (\nabla\xx)_1 \cos \ttheta +  (\nabla\xx)_2 \sin\ttheta) + \|\nabla\xx\|^2\\
& = & \left(\rho - \gamma \langle\overline{\nabla\f},\overline{\nabla\xx}\rangle\right)^2 + \gamma^2\left(1-\langle\overline{\nabla\f},\overline{\nabla\xx}\rangle^2\right)
\end{array}
\ee
We are now interested in the marginal of \eqref{eq-polar-polar} with respect to  $\rho$, \ie
\be\label{eq-marginal-alpha}
p(\ttheta|\nabla\xx) = \int_0^{\infty} p(\rho,\ttheta| \nabla\xx)\, d\rho.
\ee
where we can isolate the factor that does not depend on $\rho$, 
\be\label{eq-marginal-alpha2}
p(\ttheta|\nabla\xx)= \frac{1}{\sqrt{2\pi\epsilon^2}} e^{-\frac{1}{2\epsilon^2}\gamma^2\left(1-\langle\overline{\nabla\f},\overline{\nabla\xx}\rangle^2\right) } 
\underbrace{\int_0^{\infty} \frac{1}{\sqrt{2\pi\epsilon^2}} e^{-\frac{1}{2\epsilon^2}\left(\rho - \gamma \langle\overline{\nabla\f},\overline{\nabla\xx}\rangle\right)^2}\, \rho d\rho}_M.
\ee
The bracketed term $M$ is the integral on the interval $[0,\infty)$ of  a Gaussian density with mean $m \doteq \gamma \langle\overline{\nabla\f},\overline{\nabla\xx}\rangle = \cos(\angle \nabla \f - \angle \nabla \xx)  \| \nabla \xx \| $ and variance $\epsilon^2$; it can be rewritten, using the change of variable $\xi \doteq  (\rho-m)/\epsilon$, as 
$
%\int_0^{\infty} \frac{1}{\sqrt{2\pi\epsilon^2}} e^{-\frac{1}{2\epsilon^2}\left(\rho - \gamma \langle\overline{\nabla\f},\overline{\nabla\xx}\rangle\right)^2}\, \rho d\rho=
\int_{-m/\epsilon}^\infty  \frac{1}{\sqrt{2\pi}} e^{\frac{1}{2} \xi^2} \left(\epsilon \xi + m\right) d\xi
$
%Defining $\Psi(a) \doteq \frac{1}{\sqrt{2\pi}} \int_{-\infty}^{a}  e^{-\frac{1}{2}\tau^2} \,d\tau$,  recalling that $\frac{d}{d\xi} e^{-\frac{1}{2} \xi^2} = -\xi e^{-\frac{1}{2} \xi^2}$ and 
which can be integrated by parts to yield 
$$
%\int_0^{\infty} \frac{1}{\sqrt{2\pi\epsilon^2}} e^{-\frac{1}{2\epsilon^2}\left(\rho - \gamma \langle\overline{\nabla\f},\overline{\nabla\xx}\rangle\right)^2}\, \rho d\rho 
M = \frac{\epsilon e^{-\frac{1}{2} \frac{m^2}{\epsilon^2}}}{\sqrt{2\pi}} + m \left(1-\Psi\left(-\frac{m}{\epsilon}\right)\right) %;  \quad\quad  m=\gamma \langle\overline{\nabla\f},\overline{\nabla\xx}\rangle
$$
and therefore 
\be
p(\ttheta | \nabla \xx) =  \frac{1}{\sqrt{2\pi\epsilon^2}} \exp\left(
-\frac{\| \nabla \xx \|^2 - m^2}{2\epsilon^2} 
\right) M
%\underbrace{\left(
%\frac{\epsilon e^{-\frac{1}{2} \frac{m^2}{\epsilon^2}}}{\sqrt{2\pi}} + m \left(1-\Psi\left(-\frac{m}{\epsilon}\right)\right)
%\right)}_{M}
\ee
which, once written explicitly in terms of $\xx$ and $\f$, yields \eqref{eq-contr-inv}-\eqref{eq-M}. 
\end{proof}

{\bf Claim \ref{thm-ideal}}
\begin{proof}
Since $p_{\M,G}(\f,\xx^t)$ is sufficient for $\M$, and it factorizes into $\phi_{\M,G}(\f)\phi_\M(\xx^t)$, then $\phi_\M(\xx^t)$ is sufficient of $\xx^t$ for $\M$. By the factorization theorem (Theorem 3.1 of \cite{pawitan}), there exist functions $f_\M$ and $\psi$ such that $\phi_\M(\xx^t) \propto f_\M(\psi(\xx^t))$, \ie the likelihood depends on the data only through the function $\psi(\cdot)$. This latter function is what is more commonly known as the sufficient statistic, which in particular has the property that $p(\M| \xx^t) = p(\M | \psi(\xx^t))$. However, if $\phi_\M$ is sufficient for $\M$, it is also sufficient for future data generated from $\M$. Formally, 
\be
p_G(\f | \xx^t) = \int p_G(\f | \xx^t, \M)p(\M | \xx^t)dP(\M) = \int p_G(\f | \M)p(\M | \psi(\xx^t)) dP(\M) = p_G(\f | \psi(\xx^t))
\ee
which shows that $\psi$ minimizes the uncertainty of $\f$ for any $g\in G$ since the right-hand-side is $G$-invariant. 
The right-hand side above is the {\em predictive likelihood}\cutForReview{\footnote{Definition 2 of \cite{hinkley1979predictive} with $Y = \xx^t, Z = \f, S = \phi(\xx^t), T = \phi_G(\f)$ and $R = \phi(\xx^t, \f)$, up to a factor $\frac{p(\phi(\xx^t, \f))}{p(\phi(\xx^t))}$.}}  \cite{hinkley1979predictive}, which must therefore be proportional to $\tilde f_\f(\psi(\xx^t))$ for some $\tilde f$ {\em and the same} $\psi$, also by the factorization theorem. 
\end{proof}

{\bf Claim \ref{claim-1stlayer}}
\begin{figure}[htb]
\begin{center}
\includegraphics[width=.25\textwidth]{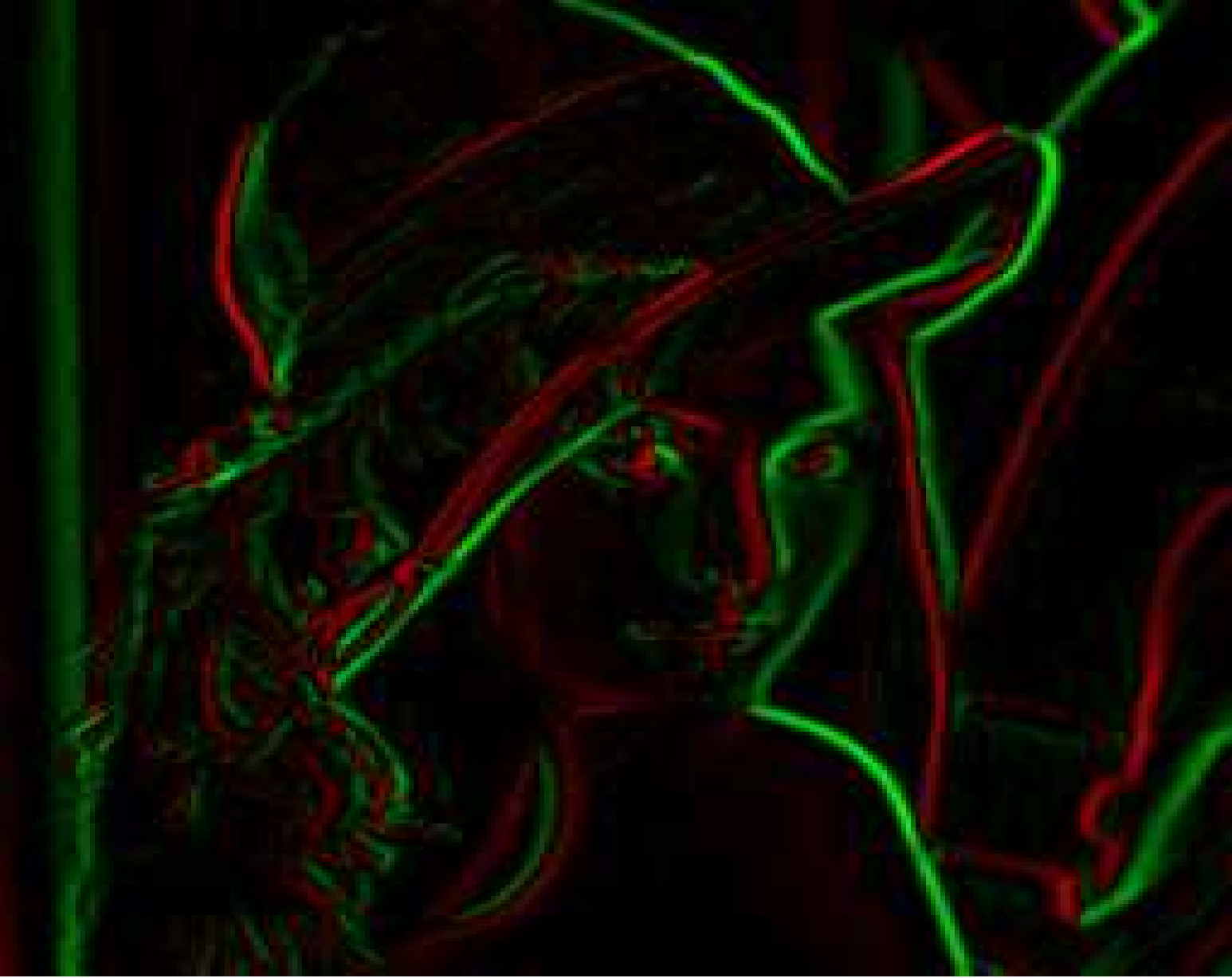}
\end{center}
\caption{\sl \small  $D_+(0)$ (green) and $D_-(0)$, the positive and negative responses to a gradient filter in the horizontal direction. The black region is their complement, which separates them.}
\label{fig-dPlusMinuse}
\end{figure}

\begin{figure}[htb]
\begin{center}
\includegraphics[width=.25\textwidth]{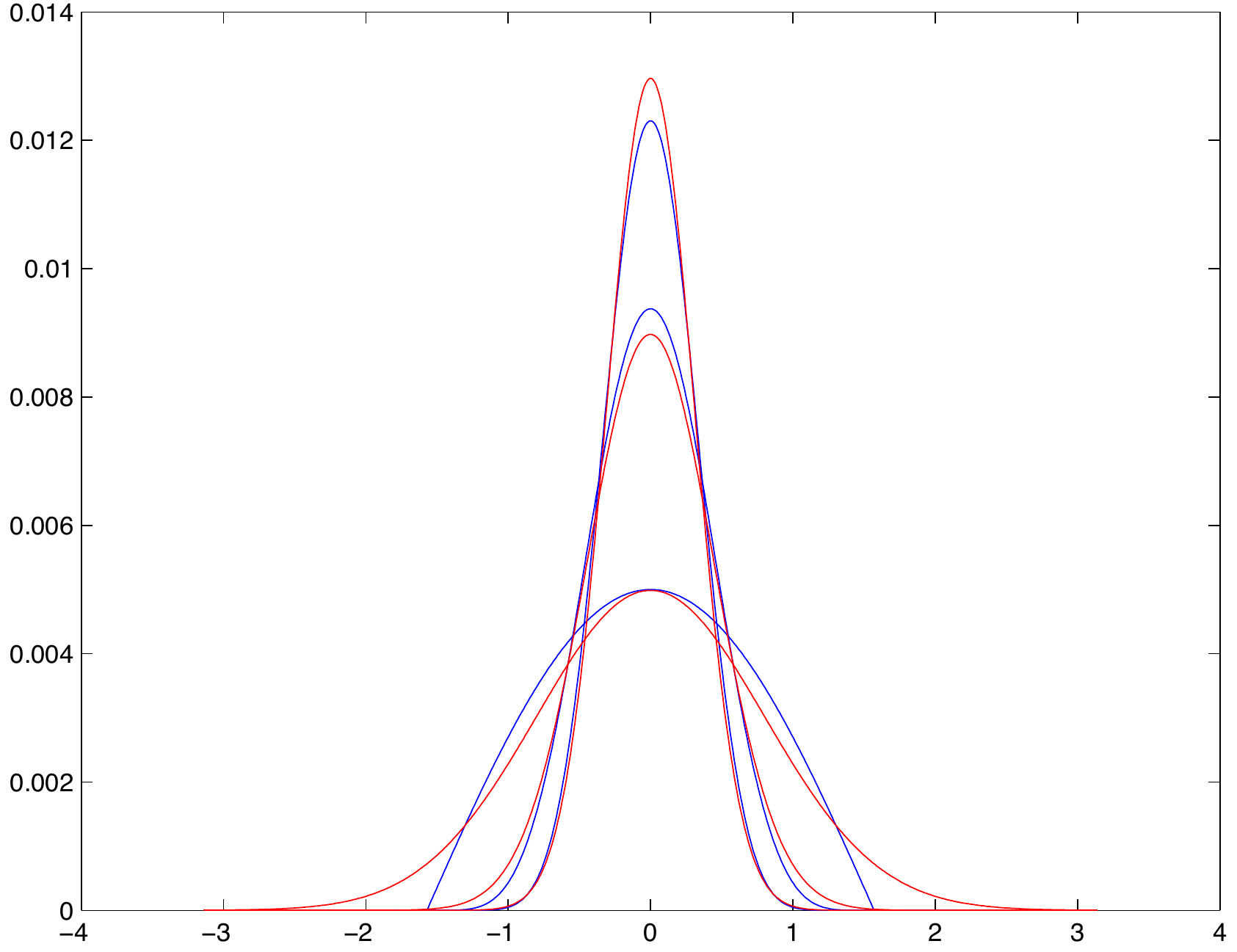}
\end{center}
\caption{\sl \small  Rectified cosine (blue) and its powers, compared to a Gaussian kernel (red). While the two are distinctly different for $\epsilon = 1$, as the power/dispersion decreases, the latter approximates the former. The plot shows $\epsilon = 1, 1/5, 1/9$ for the cosine, and $1/5, 1/9, 1/13$ for the Gaussian.}
\label{fig-cosine}
\end{figure}

\begin{proof}
(Sketch) The integral in \eqref{eq-norelu} can be split into 3 components, one of which omitted, leaving the positive component integrated on $D_+$, the negative component on $D_-$. If the distance between these two is greater than $\sigma$, however, the components are disjoint, so for each $(u,v)$ and $\alpha$, only the the positive or the negative component are non-zero, and since ${\cal N}$ and $\| \nabla \xx \|$ are both positive, and the sign is constant, rectification inside or outside the integral is equivalent. When $\sigma > d$ there is an error in the approximation, that can be bounded as a function of $\sigma$, the minimum distance and the maximum gradient component.
\end{proof}
A more general kernel could be considered, with a parameter $\epsilon$ that controls the decay, or width, of the kernel, $\kappa_\epsilon(\alpha)$. For instance, $\kappa_\epsilon(\alpha) = \kappa(\alpha)^{\frac{1}{\epsilon}}$, with the default value being $\epsilon = 1$. An alternative is to define $\kappa$ to be an angular Gaussian with dispersion parameter $\epsilon$, which is constrained to be positive and therefore does not need rectification.  Although the angular Gaussian is quite different from the cosine kernel for $\epsilon = 1$, it approximates it as $\epsilon$ decreases (Fig. \ref{fig-cosine}).
A corollary of the above is that {\em the visible layer of a CNN computes the SAL Likelihood of the first hidden layer.}

The interpretation of SIFT as a likelihood function given the test image $\f$ can be confusing, as ordinarily it is interpreted as a ``feature vector'' associated to the training image $\xx$, and compared with other feature vectors using the Euclidean distance. In a likelihood interpretation, $\xx$ is used to compute the likelihood function, and $\f$ is used to evaluate it. So, there is no descriptor built for $\f$. The same interpretational difference applies to convolutional architectures. If interpreted as a likelihood, which would require generative learning, one would compute the likelihood of different hypotheses given the test data. Instead, currently the test data is fed to the network just as training data were, thus generating features maps, that are then compared (discriminatively) by a classifier.

\end{document}